\documentclass{article}

\usepackage{arxiv}

\usepackage[utf8]{inputenc} 
\usepackage[T1]{fontenc}    
\usepackage[hidelinks]{hyperref}       
\usepackage{url}            
\usepackage{booktabs}       
\usepackage{amsfonts}       
\usepackage{nicefrac}       
\usepackage{microtype}      
\usepackage{lipsum}
\usepackage{graphicx}
\graphicspath{ {./figures/} }

\usepackage{boldline} 
\usepackage{subcaption}
\usepackage{multirow}
\usepackage{multicol}
\usepackage{bm}
\usepackage{amsmath}
\usepackage[font=small,labelfont=bf,figurename=Figure]{caption} 
\usepackage[capitalize,nameinlink]{cleveref}
\crefname{section}{Section}{Sections}
\crefname{subsection}{Subsection}{Subsections}

\title{Enhancing Physics-Informed Neural Networks with Domain-aware Fourier Features:
Towards Improved Performance and Interpretable Results}

\author{
 Alberto {Miño Calero} \\
  Department of Engineering Cybernetics\\ 
  NTNU, Trondheim, Norway\\
  \texttt{ziq2@pitt.edu} \\
   \And
 Luis Salamanca \\
  Swiss Data Science Center (SDSC)\\
  ETH Z{\"u}rich, Z{\"u}rich, Switzerland \\
  \texttt{luis.salamanca@sdsc.ethz.ch} \\
  \And
 Konstantinos E. Tatsis \\
  Swiss Data Science Center (SDSC)\\
  ETH Z{\"u}rich, Z{\"u}rich, Switzerland \\
  \texttt{konstantinos.tatsis@sdsc.ethz.ch} \\
}

\begin{document}
\maketitle
\begin{abstract}
Physics-Informed Neural Networks incorporate underlying physics into neural networks by embedding partial differential equations into their loss function. Despite their success in learning underlying physics, these models remain difficult to train and interpret. In this work, a novel approach is proposed with Domain-aware Fourier Features for the positional encoding of the input space. These features encapsulate domain-specific characteristics, such as the geometry and boundary conditions, and, unlike Random Fourier Features, eliminate the need for explicit boundary condition loss terms and loss balancing schemes, while simplifying the optimization process and reducing the computational cost of training. We further develop a Layer-wise Relevance Propagation-based explainability framework tailored to Physics-Informed Neural Networks, enabling extraction of relevance attribution scores for the input space. It is demonstrated that the models with Domain-aware Fourier Features achieve orders-of-magnitude lower errors and converge faster compared to vanilla and Random Fourier Features-based ones. Furthermore, the explainability analysis reveals that the proposed approach leads to more physically consistent feature attributions, in contrast to more scattered and less physics-relevant patterns displayed by others. These results demonstrate that our approach not only enhances Physics-Informed Neural Networks' accuracy and efficiency but also improves interpretability, laying groundwork for more robust and informative physics-informed learning.
\end{abstract}


\section{Introduction}
\label{sec:introduction}

The increased availability of data and computational resources, coupled with the development of pattern recognition techniques \cite{sarker_machine_2021}, has positioned deep learning approaches as central to the evolution of predictive models \cite{sarker_deep_2021}. Thanks to their flexibility and representational power \cite{lu_universal_2020}, these models can tackle the most complex problems with their typically over-parametrized and challenging to interpret architectures, yielding impressive accuracy. However, in many scenarios, the data acquisition still poses a challenge. Additionally, deep learning models struggle with a variety of complex tasks, suffering from various issues during training, including local minima, convergence problems, abnormal gradient behavior, overfitting, and poor extrapolation capabilities. 

Typical feed-forward neural networks do not embed prior knowledge into their training. However, we can find models tailored to specific tasks. Convolutional neural networks \cite{lecun_convolutional_1998}, for computer vision, intrinsically make the networks exploit spatial locality, reduce the dimensionality, and understand visual patterns such as symmetries. Transformers \cite{vaswani_attention_2017}, which are utilized in natural language processing, excel due to their attention mechanism, permutation-invariant nature, and positional embeddings, enabling them to handle contextual information effectively and understand natural language, resulting in remarkable performance in various downstream tasks, such as classification and generative solutions. While these approaches do not embed prior knowledge implicitly, their architectures are designed to handle specific types of data better. 

Deep learning models that incorporate physics-based knowledge during training have been highlighted in several recent literature reviews because of their rapid development and growing importance in computational mechanics. In this context, there is a range of approaches, including conventional data-driven methods \cite{Bai2025ttf}, as well as Physics-Guided, Physics-Informed, and Physics-Encoded Neural Networks, and Neural Operators \cite{Faroughi2024pgp}, that have contributed to advancing computational fluid and solid mechanics. However, these works also reveal persistent challenges related to robustness, generalization, and computational efficiency.

Among these frameworks, Physics-Informed Neural Networks (PINNs) \cite{raissi_physics-informed_2019, raissi_physics_2017} are notably relevant due to their capability to learn from limited labeled data and enforce physical consistency through governing equations. Moreover, recent survey work \cite{Hu2024pin} highlights the relevance of PINNs in computational solid mechanics, with developments ranging from methodological aspects to applications in constitutive characterization, inverse identification, and problems related to damage and fracture in structures and materials.

PINNs are a type of deep learning architecture that embeds prior knowledge of the systems they model by making use of automatic differentiation \cite{baydin_automatic_2017}. They are trained with physics knowledge embedded in the loss function. In particular, PINNs include one or several Partial Differential Equations (PDEs) in the loss function, which serve as regularization terms that ensure that the models learn to solve the task, minimizing the residual error of the known physics. 

Despite the PDEs' guidance, PINNs are still black-box models that do not offer an understanding of their inference process and biases, and how the problems are handled. However, addressing PINNs through the lenses of eXplainable Artificial Intelligence (XAI) remains a topic significantly overlooked. A likely cause for this gap is the perception that PINNs are more interpretable due to the physics knowledge used for their training. However, the embedded PDEs are just a guidance to minimize the residual errors, but compliance is not guaranteed. The loss function does not explain how the model is handling the information, or if it is simulating the dynamics of the process through an approximation that, despite possibly yielding good metrics, does not handle the inputs of the equations in compliance with the known dynamics. 

Analyzing black-box models is the main goal of XAI methods \cite{mersha_explainable_2024}, which provide explanations that approximate the inner workings of models in a variety of ways \cite{chamola_review_2023,antamis_interpretability_2024}. Many XAI methods focus on explaining predictions through local input feature contribution, also known as attribution scores. However, when the inputs are merely spatial or temporal coordinates, which happens often in problems solved with PINNs, the analysis becomes more challenging because simple feature attribution scores (also known as input contributions) assigned to each coordinate lack explanatory power. 

To the best of our knowledge, existing XAI-oriented work on PINNs is still relatively sparse. For instance, an XAI-based analysis with classical methods in forecasting settings \cite{Inam2025ana}, which uses Shapley Additive Explanations (SHAP) \cite{lundberg_unified_2017} and Local Interpretable Model-agnostic Explanations (LIME) \cite{ribeiro_why_2016} in particular, but the inputs include environmental features rather than spatial coordinates. Another study \cite{kim_physics-informed_2025} applies Layer-wise Relevance Propagation (LRP) \cite{bach_pixel-wise_2015} to a remaining useful life prediction model. Nevertheless, the model is not a PINN but a multi-scale convolutional network that uses a prior physics-informed feature transformation process, and it does not contain governing equations in its loss term. An example of a more exploratory approach is \cite{Naujoks2026lif}, where the importance of training data is measured to guide adaptive resampling of collocation points, but in this case, no interpretability of the learned physical representations is accomplished.

As mentioned before, many challenges are associated with training PINNs, some of which are related to their multi-objective nature, where the different PDE terms that contribute to the loss function often drive the learning space in different directions. Issues such as gradient stiffness \cite{Wang2021} or bias towards low-frequency solutions \cite{Wang2020a} impair the capacity of these models to learn the complex features that can often be found in PDE problems. 

To tackle the described problems, this work first proposes a novel type of Fourier features called Domain-aware Fourier Features (DaFFs) that allow the embedding of information about the geometry. With DaFFs, the models naturally satisfy a certain type of boundary condition where the solution is equal to a constant zero value. This gives an advantage over Random Fourier Features (RFFs), since the loss terms for the boundary conditions can be removed, simplifying the training of the PINNs from a multi-objective setup to a single-objective one: fitting the PDE. We test our approach with two numerical examples and provide a performance comparison between vanilla PINNs, PINN-RFFs, and PINN-DaFFs. Second, we look into the models' behavior via XAI and provide an extensive analysis on the matter, showing that it is possible to obtain knowledge from the model even when the inputs only consist of spatial coordinates. We also showcase the usage of XAI to select features when dealing with transformations such as those applied by Fourier transforms and randomly sampled parameters.

\section{Problem description}
\label{sec:problem-description}

This section describes the problem of PINNs. Without loss of generality, we consider PDEs of the following form
\begin{subequations}
    \begin{align}
        \mathcal{N}[\mathbf{u}](\mathbf{x}, t) & = \mathbf{q}(\mathbf{x}, t),\quad \mathbf{x}\in\Omega, \quad t\in T
        \label{eq:pde-eq}\\[2mm]
        \mathcal{B}[\mathbf{u}](\mathbf{x}, t) & = \mathbf{0},\quad \mathbf{x}\in\partial\Omega, \quad t \in T
        \label{eq:pde-bc}\\[2mm]
        \mathcal{C}[\mathbf{u}](\mathbf{x}, t) & = \mathbf{0}, \quad \mathbf{x}\in\Omega, \quad t=t_0
        \label{eq:pde-ic}
    \end{align}
    \label{eq:pde}
\end{subequations}
\noindent
where $\mathcal{N}[\cdot]$ is a differential operator with respect to the spatial and temporal coordinates, $\mathbf{x}$ and $t$ respectively. Similarly, $\mathcal{B}[\cdot]$ denotes the operator associated with the boundary conditions, which can be of periodic type, Dirichlet, Neumann, or Robin. In general, initial conditions can be treated as a special type of boundary conditions, with $\mathcal{B}[\cdot]$ acting on both spatial and temporal coordinates. However, these are herein separated, and the initial conditions are described by the operator $\mathcal{C}[\cdot]$, \cref{eq:pde-ic}. Lastly, $\mathbf{u}: \Omega\times T \rightarrow \mathbb{R}^{n}$ is the sought-after solution, governed by \cref{eq:pde}, with $\Omega$ and $T$ denoting the spatial and temporal domains, while $\partial\Omega$ indicates the domain boundary.

\subsection{Physics-Informed Neural Networks}
\label{sub:pinns}

In the context of PINNs, the solution of \cref{eq:pde} is learned by a Neural Network (NN), which is trained to minimize the residuals associated with all three PDE equations. A vanilla implementation of PINNs generally relies on a fully-connected feed-forward NN, comprised of $L$ hidden layers and defined by the trainable parameter vector $\bm{\theta}$. It receives as input the spatial and temporal coordinates, $\mathbf{x}$ and $t$. Consequently, the network layers operating on the input can be described recursively for $h=1, \ldots, L$, as follows
\begin{subequations}
    \begin{align}
        \mathbf{g}^{(0)}(\mathbf{x}, t;\bm{\theta}) & = \mathrm{vec}\left([\mathbf{x}, t]\right)\label{eq:NN-hidden-layers-1}\\[2mm]
        \mathbf{f}^{(h)}(\mathbf{x}, t;\bm{\theta}) & = \frac{1}{\sqrt{d_{h-1}}}\mathbf{W}^{(h-1)}\mathbf{g}^{(h-1)}(\mathbf{x}, t;\bm{\theta}) + \beta^{(h-1)}\\
        \mathbf{g}^{(h)}(\mathbf{x}, t;\bm{\theta}) & = \sigma\left(\mathbf{f}^{(h)}(\mathbf{x}, t;\bm{\theta})\right)
    \end{align}
    \label{eq:NN-hidden-layers}
\end{subequations}
where $\mathbf{W}^{(h)}\in\mathbb{R}^{d_{h+1}\times d_h}$ and $\beta^{(h)}\in\mathbb{R}^{d_{h+1}}$ denote the weight matrix and the bias vector of the $h$-th hidden layer respectively, $d_{h}$ indicates the corresponding dimensions of the $h$-th layer and $\sigma:\mathbb{R}\to\mathbb{R}$ denotes the point-wise activation function. The predicted output at the last layer of the model $\hat{\mathbf{u}}(\mathbf{x},t;\bm{\theta})$ is obtained as
\begin{equation} 
    \hat{\mathbf{u}}(\mathbf{x}, t;\bm{\theta}) = \mathbf{g}^{(L+1)}(\mathbf{x}, t;\bm{\theta}) = 
    \frac{1}{\sqrt{d_{L}}}\mathbf{W}^{(L)}\mathbf{g}^{(L)}(\mathbf{x}, t;\bm{\theta}) + 
    \beta^{(L)}
    \label{eq:NN-output}
\end{equation}
with parameter vector $\bm{\theta} = \left[\mathbf{W}^{(0)},\ \beta^{(0)},\  \ldots, \mathbf{W}^{(L)},\beta ^{(L)}\right]$.

During training, the objective is to jointly satisfy the PDE terms described by \cref{eq:pde}, as enforced by the following loss function
\begin{equation}
    L(\bm{\theta}) = \lambda_{\mathrm{r}}L_{\mathrm{r}}(\bm{\theta}) + \lambda_{\mathrm{b}}L_{\mathrm{b}}(\bm{\theta}) + \lambda_{\mathrm{c}}L_{\mathrm{c}}(\bm{\theta})
    \label{eq:loss}
\end{equation}
where $L_{\mathrm{r}}$ denotes the loss term associated with the PDE itself, while $L_{\mathrm{b}}$ and $L_{\mathrm{c}}$ are related to the boundary and initial conditions, respectively. The weighting terms $\lambda_{\mathrm{r}}$, $\lambda_{\mathrm{b}}$ and $\lambda_{\mathrm{c}}$ are used to balance the contribution of each component. For the evaluation of the loss terms, a set of spatial and temporal collocation points is considered, resulting in the following expressions
\begin{subequations}
    \begin{align}
        L_{\mathrm{r}}(\bm{\theta}) & = \frac{1}{N_{\mathrm{r}}N_{\mathrm{t}}}\sum_{k=1}^{N_{\mathrm{t}}}\sum_{j=1}^{N_{\mathrm{r}}}
        \left|\mathcal{N}[\hat{\mathbf{u}}](\mathbf{x}_{\mathrm{r}}^j, t^k;\bm{\theta}) - \mathbf{q}(\mathbf{x}_{\mathrm{r}}^j, t^k)\right|^2\\
        L_{\mathrm{b}}(\bm{\theta}) & = \frac{1}{N_{\mathrm{b}}N_{\mathrm{t}}}\sum_{k=1}^{N_{\mathrm{t}}}\sum_{j=1}^{N_{\mathrm{b}}}
        \left|\mathcal{B}[\hat{\mathbf{u}}](\mathbf{x}_{\mathrm{b}}^j, t^k;\bm{\theta}) \right|^2\\
        L_{\mathrm{c}}(\bm{\theta}) & = \frac{1}{N_{\mathrm{r}}}\sum_{j=1}^{N_{\mathrm{r}}}
        \left|\mathcal{C}[\hat{\mathbf{u}}](\mathbf{x}_{\mathrm{r}}^j, t_0;\bm{\theta})\right|^2
    \end{align}
\end{subequations}
where $N_{\mathrm{r}}$, $N_{\mathrm{b}}$ and $N_{\mathrm{t}}$ denote the number of collocation points associated with $\Omega$, $\partial\Omega$ and $T$ respectively.

One of the major shortcomings encountered during the training of PINNs is the stiffness in the gradient flow dynamics, characterized by the convergence rate discrepancy among the different components of the loss function in \cref{eq:loss}. This limitation has been extensively explored in \cite{Wang2021}, and a loss-balancing solution using gradient statistics from the training has been proposed. Several alternatives \cite{Chen2017,Heydari2019,Bischof2021} have also been proposed to tackle this issue, employing various adaptive strategies designed to balance the loss terms. These approaches aim to enhance trainability and predictive accuracy while also decreasing the computational overhead. 

Additionally, PINNs, and specifically their learning space, are strongly limited by the architecture. This issue has been studied \cite{Jacot2018,Yang2019} from a Neural Tangent Kernel (NTK) point of view, which is central to describing the generalization characteristics of NNs. From this perspective, the learning dynamics of PINNs are dictated by their NTK, which essentially governs how the model explores the function space during training. As such, the NTK of vanilla PINNs, built on fully-connected NNs, introduces a bias towards smooth, low-frequency solutions, often restricting the model’s ability to capture complex or high-frequency features, which are inherent in many PDE problems \cite{Wang2020a}. This limitation can be mitigated by increasing the depth or width of the network to enrich the NTK, or by augmenting the input space with RFFs, which expand the representational capacity and enable the network to better approximate functions with high-frequency variations \cite{Wang2020b}.

\subsection{Random Fourier Features}
\label{sub:fourier-features}

The spectral bias of vanilla PINNs, characterized by the low-frequency approximations associated with the limiting eigen-directions of the NTK \cite{Wang2020a}, can be significantly improved with the use of RFFs as coordinate embeddings of the input. This simple transformation enables the learning of high-frequency solutions \cite{Tancik2020}, thus increasing the effectiveness of PINNs across different tasks. The positional encoding of the input coordinates with RFFs can be expressed as
\begin{equation}
    \phi_j(\mathbf{x}) = 
    \begin{bmatrix}
        \cos(\mathbf{b}_j\mathbf{x})\\[1mm]
        \sin(\mathbf{b}_j\mathbf{x})
    \end{bmatrix}
    \label{eq:random_FF}
\end{equation}
where each entry of $\mathbf{b}_j\in\mathbb{R}^{1\times d}$ is obtained from a zero-mean Gaussian distribution, so that $\mathbf{b}_{j,k}~\sim\mathcal{N}(0, \sigma_j^2)$. The variance $\sigma_j^2$ is a user-defined hyperparameter, which specifies the frequency content of the input encoding, while $d$ denotes the dimensionality of the input vector and $k = 1,\dots,d $ indexes the components of $\mathbf{b}_{j}$.

Despite the improved expressiveness of the model, achieved without the need to make the network deeper, there are shortcomings associated with RFFs. Namely, the number of Fourier features and the choice of $\mathbf{b}_j$ coefficients, which are regulated by the corresponding variance terms, play a significant role in the expressiveness of the model. Moreover, RFFs introduce a non-trainable randomness, which can affect reproducibility. Lastly, although the spectral bias is corrected, the stochastic nature of the features does not improve the interpretability of the models since the output energy is distributed throughout the entire input space.

\subsection{Domain-aware Fourier Features}
\label{sub:domain-aware-features}

To address the implicit shortcomings of RFFs, a set of DaFFs is herein proposed. The DaFFs are derived from the provided domain equation, subject to the corresponding boundary and initial conditions, lifting the non-trainable randomness of RFFs. Consequently, the loss function is simplified to a single term, the one associated with the PDE residuals, which does not require adaptive loss-balancing, ultimately easing the training process.

The idea of DaFFs stems from the approximation of stationary kernels by means of an eigenvalue decomposition. Without loss of generality, the idea is herein presented for Dirichlet boundary conditions, and it can be easily extended to Neumann- and Robin-type conditions, as will be demonstrated in \cref{sec:numerical-examples}. Therefore, the eigenvalue decomposition is applied to the Laplace operator on the domain $\Omega$, subject to the boundary conditions imposed by \cref{eq:pde-bc}, described by the following set of equations
\begin{subequations}
    \begin{alignat}{2}
        -\nabla^2 \phi_j(\mathbf{x}) & = \lambda_j \phi_j(\mathbf{x}), \quad &&\mathbf{x}\in\Omega\label{eq:Laplace-operator-1}\\[1mm]
        \mathbf{h}\left(\phi_j (\mathbf{x})\right) & = 0, \quad &&\mathbf{x}\in\partial\Omega\label{eq:Laplace-operator-2}
    \end{alignat}
    \label{eq:Laplace-operator}
\end{subequations}
with $\lambda_j$ and $\phi_j(\mathbf{x})$ denoting the $j$-th eigenvalue and eigenfunction of the Laplace operator, respectively, while $\mathbf{h}:\partial\Omega \to \mathbb{R}^n$ denotes the function of boundary conditions, in accordance with the PDE, as defined in \cref{eq:pde}. 

For certain types of domains, such as rectangles, circles, and spheres, the eigendecomposition can be computed in closed form \cite{Solin2014}.

For more complex domains, the eigendecomposition of the Laplace operator on the domain $\Omega$ of the PDE can be computed numerically, using, for instance, Finite Difference (FD) approximation. Within this context, if we consider a domain in the 2D space, this is first discretized using a grid of $n\times n$ points and the Laplacian is approximated using a $k$-point FD scheme with step size $h$. These steps enable the conversion of \cref{eq:Laplace-operator} into the following discretized form

\begin{equation}
    \mathbf{L}\,\bm{\phi}_j = \lambda_j \bm{\phi}_j
    \label{eq:discretized-Laplace-operator}
\end{equation}

\noindent
which is defined by means of the Laplacian matrix $\mathbf{L}\in\mathbb{R}^{N\times N}$, where $N=n^2$. Moreover, $\bm{\phi}_j\in\mathbb{R}^{N}$ denotes the discretized eigenfunction evaluated at the grid points, so that the $i$-th entry of $\bm{\phi}_{j}$, namely $\bm{\phi}_{ji}$, denotes the $j$-th eigenfunction evaluated at the $i$-th grid point $\phi_j(\mathbf{x}_i)$, with $\mathbf{x}_i=(x_i,y_i)$ for $i=1,\ldots,N$. Due to the sparsity of the Laplacian matrix, the first few eigenvalues and eigenvectors can be easily extracted using iterative methods. Although the extraction of DaFFs is presented for homogeneous boundary conditions, it can be extended to problems with non-homogeneous conditions at the boundaries. In this scenario, the solution for \cref{eq:Laplace-operator} is decomposed into a part that solves the homogeneous case, relying on the solution of \cref{eq:discretized-Laplace-operator}, and a second part that satisfies the boundary condition, learned separately using an approach similar to the ones presented in \cite{Berg2018,Kraus2024}.

It should be noted that the Laplace equation can accommodate any type of boundary conditions, namely homogeneous or non-homogeneous, Dirichlet, Neumann, or Robin, as long as these can be expressed in the form of \cref{eq:Laplace-operator-1}. However, the use of DaFFs is not motivated only by the treatment of boundary conditions, which are satisfied by design, but it further offers an efficient inductive bias towards the improvement of training, as well as expressiveness and interpretability of the model. Moreover, the use of DaFFs is not exclusive to the model architecture presented herein, but can be tailored to all different formulations that aim to address the imposition of boundary conditions. Namely, the DaFFs can be used for the training of hard-constraint PINNs \cite{lu_physics-informed_2021} or PINNs that aim for the exact imposition of boundary conditions \cite{sukumar_exact_2022}. In the context of these formulations, the solution of the PDE itself is separated from the boundary conditions, which results in two models: one learning the solution of the PDE without the boundary conditions and a second one learning the solution at the boundaries. As such, DaFFs can be used as an additional feature in order to significantly increase the expressiveness of such models.

Since the DaFFs are designed as spectral components that satisfy the homogeneous boundary conditions of the Laplace operator eigenvalue problem in \cref{eq:Laplace-operator}, each DaFF must yield a zero value at the boundaries and the initial point in time. This implies that the DaFF values at the boundaries are propagated through the network, which happens naturally if the biases are removed from the model, thus allowing \cref{eq:pde-bc,eq:pde-ic} to be intrinsically satisfied. The propagation of a zero value through the network is enabled by most activation functions used nowadays, such as the identity function, linear function, tanh, ReLU, Leaky ReLU, GELU, and ELU. Without $\beta^{(h)}$, now the model results in
\begin{subequations}
    \begin{align}
        \mathbf{f}^{(h)}(\mathbf{x}, t;\bm{\theta}) & = \frac{1}{\sqrt{d_{h-1}}}\mathbf{W}^{(h-1)}\mathbf{g}^{(h-1)}(\mathbf{x}, t;\bm{\theta})\\[2mm]
        \mathbf{g}^{(h)}(\mathbf{x}, t;\bm{\theta}) & = \sigma\left(\mathbf{f}^{(h)}(\mathbf{x}, t;\bm{\theta})\right)
    \end{align}
\end{subequations}
\noindent
where $\mathbf{g}^{(0)}(\mathbf{x}, t;\bm{\theta})$ is given by \cref{eq:NN-hidden-layers-1}.

In the case of simple rectangular domains, where $\Omega= [0,a] \times [0, b]$ and homogeneous Dirichlet conditions are assumed at the boundary, the DaFFs can be derived analytically from the eigenvectors of \cref{eq:Laplace-operator} leading to:
\begin{equation}
    \phi_j(\mathbf{x}) = 
    \sin\left(\frac{m_j\pi x}{a}\right)
    \sin\left(\frac{n_j\pi y}{b}\right), \quad\forall \mathbf{x}:=(x,y) \in \Omega
\end{equation}
along with the corresponding eigenvalues
\begin{equation}
    \lambda_j = \pi^2
    \left(
    \left(\frac{m_j}{a}\right)^2+
    \left(\frac{n_j}{b}\right)^2
    \right)
\end{equation}
where $m_j$ and $n_j$ are integers denoting the mode numbers in $x$ and $y$ directions, respectively, which control the frequency content of the model output. Therefore, higher mode numbers indicate higher frequency content of the PINN solution.

Furthermore, the use of DaFFs enables the satisfaction of Neumann and higher-order Neumann-type boundary conditions. This becomes evident from the expression of $k$-th order derivatives in $x$ and $y$ directions, defined as
\begin{subequations}
	\begin{align}
		\frac{\partial^{k}\phi_j(\mathbf{x})}{\partial{x}^k} & = 
		\left(\frac{m_j\pi}{a}\right)^k
		\sin\left(\frac{m_j\pi x}{a}+\frac{k\pi}{2}\right)
		\sin\left(\frac{n_j\pi y}{b}\right), 
		\quad\forall (x,y) \in \Omega\\
		\frac{\partial^{k}\phi_j(\mathbf{x})}{\partial{y}^k} & = 
		\left(\frac{n_j\pi}{b}\right)^k
		\sin\left(\frac{m_j\pi x}{a}\right)
		\sin\left(\frac{n_j\pi y}{b}+\frac{k\pi}{2}\right), 
		\quad\forall (x,y) \in \Omega
	\end{align}
\end{subequations}
which yield zero values at the boundaries $\partial\Omega$. Lastly, in order to allow the handling of more complex boundary conditions, which might consist of free boundaries or even constant values at the boundary, various permutations of the DaFFs can be utilized to increase the expressiveness of the model's inputs. This is taken into account by the eigenvalue problem of \cref{eq:Laplace-operator}. Additionally, it can be accomplished by manually combining harmonic components in $x$ and $y$ directions that contain the appropriate phase shifts. An example is shown in the following DaFFs
\begin{equation}
    \phi_j(\mathbf{x}) = 
    \sin\left(\frac{m_j\pi x}{a}\right)
    \cos\left(\frac{n_j\pi y}{b}\right), \quad\forall (x,y) \in \Omega
\end{equation}
which allow for a free boundary at $y=0$ and $y=b$, thus enabling the handling of non-homogeneous Neumann or higher-order Neumann-type boundary conditions. 

\begin{figure}[!htb]
    \centering
    \includegraphics[width=0.975\textwidth]{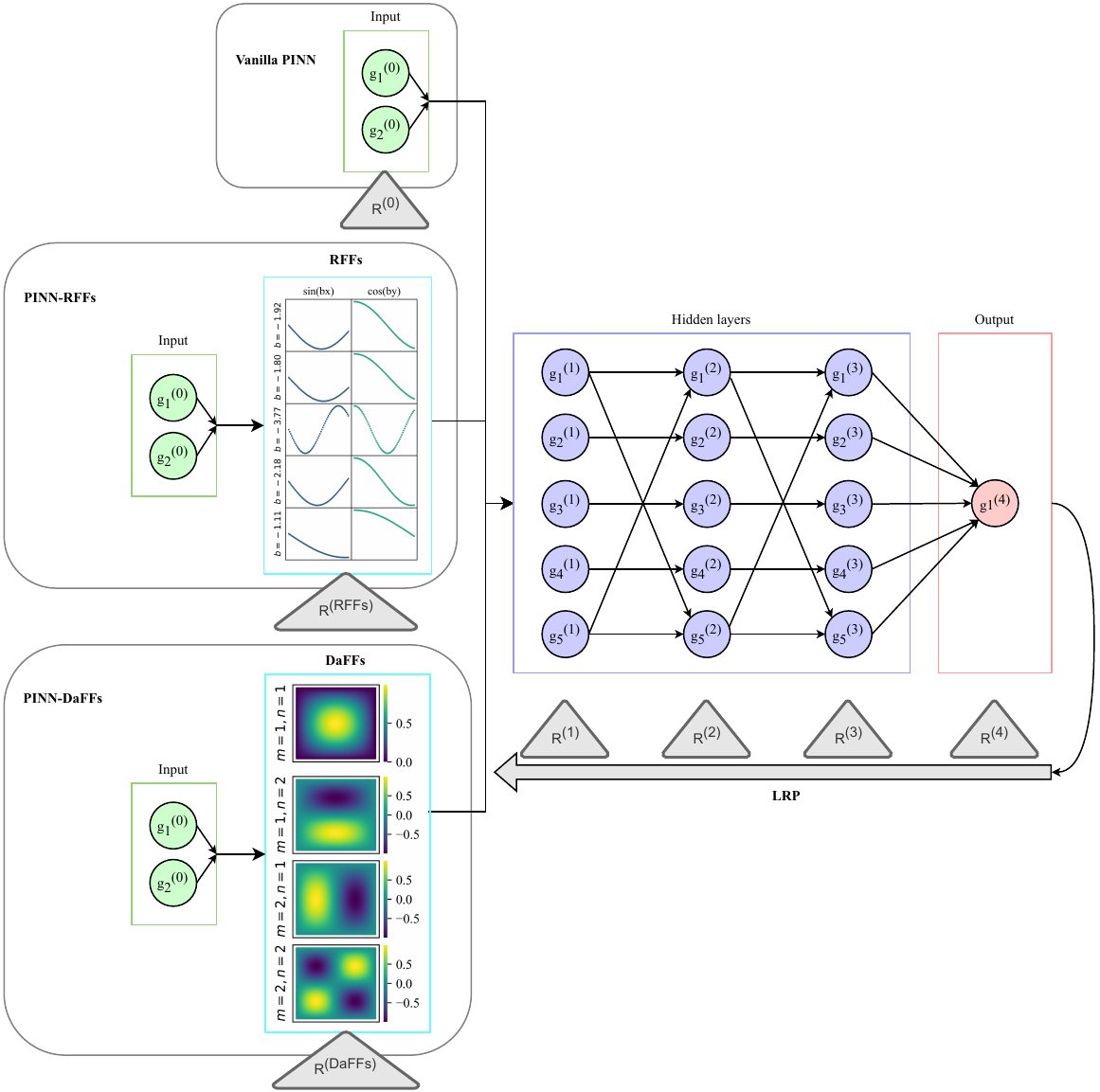}
    \caption{Schematic representation of the proposed modeling approach.}
    \label{fig:graphical-abstract}
\end{figure}

It should be noted that the presented construction, although introduced using homogeneous Dirichlet conditions, inherently supports more general boundary operators. Specifically, DaFFs are built from Laplacian eigenspaces that simultaneously control the function values and their normal derivatives at the domain boundaries. Consequently, boundary conditions involving derivative terms, as well as linear combinations of function values and derivatives, can be accommodated within the same eigensystem. The implication is that Robin boundary conditions, which combine Dirichlet's and Neumann's, are already satisfied whenever the constituent operators are enforced within the eigensystem, which lies at the core of our proposed DaFFs.

\section{Explainability}
\label{sec:explainability}

We adopt an explainability framework to highlight the advantages of the proposed modeling approach, which relies on the use of DaFFs for the encoding of the model inputs, as schematically depicted in \cref{fig:graphical-abstract}. This framework provides a quantitative assessment of the performance achieved by the proposed model, hereafter referred to as PINN-DaFFs, and a comparison with the results delivered by vanilla PINNs and PINN models combined with RFFs, denoted by PINN-RFFs. 

Explainable artificial intelligence (XAI) methodologies employ strategies such as input perturbations, gradients, and surrogate models, among others. Popular XAI perturbation-based methods include SHAP \cite{lundberg_unified_2017}, LIME \cite{ribeiro_why_2016}, and Feature ablation \cite{suresh_clinical_2017}. However, they rely on assumptions such as that, by zeroing an input, its effect on the output is nullified, or other general assumptions about how to perturb the inputs. These assumptions do not take into account geometrical constraints, which would require ad-hoc solutions for each problem. Ultimately, this can lead to poor explanations for PINNs and may also result in perturbations that break domain constraints. Alternatively, surrogate-based techniques, such as using symbolic regression via genetic programming \cite{luo_exploring_2022} or specialized decision trees \cite{zhang_interpreting_2019}, do not need to perturb the inputs, solving this issue. However, since the surrogate models do not have access to the physics, the explanations may still be poor approximations of the PINNs.

Contrary to the previous methods, gradient-based techniques \cite{ancona_gradient-based_2019} propose a solution that directly employs the PINNs to extract the explanations. Methods such as Integrated Gradients (IG) \cite{sundararajan_axiomatic_2017}, LRP \cite{bach_pixel-wise_2015}, and Deep Learning Important Features (DeepLIFT) \cite{shrikumar_learning_2019} can be directly applied without perturbing inputs or requiring the training of surrogate models. In particular, LRP constitutes a reliable and adaptable approach, with several propagation rules \cite{montavon_layer-wise_2019} that can be used to handle a large number of model features, architectures, and tasks. 

With LRP, the prediction of the model is propagated backwards, with special rules governing the contribution scores, such as the principle of conservation. This ensures that the contribution information received by each neuron is redistributed to the lower layers. The contribution is typically computed down to the input layers and used as a score that signals how much, positively or negatively, each input contributes to the prediction. The propagation of the relevance scores $R$ from a neuron $j$ in layer $h+1$ to another neuron $i$ in the previous layer $h$ is computed as follows:

\begin{align}
    R^{(h)}_{i} & = \sum_{j=1}^{d_{h}} \cfrac{z_{ij}}{\sum\limits_{i=1}^{d_{h+1}}z_{ij}}R^{(h+1)}_{j}
\end{align}

\noindent
where $z_{ij}$ models how much neuron $i$ contributes to the relevance of neuron $j$. The conservation is enforced by the denominator $\sum_{i}z_{ij}$, since $\sum_{i}R_{i}=\sum_{j}R_{j}$. The rule can be applied iteratively from the output of the model up to the input layer. Several rules exist in the literature for the computation of $z_{ij}$, being the most commonly used ones LRP-$\mathrm{\theta}$, LRP-$\mathrm{\epsilon}$, and LRP-$\mathrm{\gamma}$. LRP-$\mathrm{\theta}$ is typically unstable for noisy gradients, while LRP-$\mathrm{\gamma}$ allows control over positive and negative contributions, usually favoring positive ones. This makes sense for classification tasks, where negative contributions can be interpreted as features that most likely belong to another class. However, for regression tasks, both positive and negative contributions are equally relevant to the final value, and therefore, this rule can lead to misleading interpretations. Given these restrictions, we decide to apply LRP-$\mathrm{\epsilon}$, which offers a solution to the instability issue of LRP-$\mathrm{\theta}$ by adding a small quantity $\epsilon$ to the denominator. This results in a propagation rule where $z_{ij} = \sigma_{i}w_{ij}$, with $\sigma_i$ representing the activation of neuron $i$ and $w_{ij}$ referring to the weights that associate neuron $i$ to neuron $j$. The propagation rule is then denoted by:

\begin{equation}
    \label{eq:lrp}
    R^{(h)}_{i} = \sum_{j=1}^{d_{h}} \cfrac{\sigma_{i}w_{ij}}{\epsilon + \sum\limits_{i=1}^{d_{h+1}}\sigma_{i}w_{ij}}R^{({h+1})}_{j}
\end{equation}

LRP-$\mathrm{\epsilon}$ can be applied to fully connected layers with activation functions such as \textit{ReLU} and \textit{tanh}, with the latter being the one chosen for the models developed in this work. However, residual skip connections, also part of our model, need to be taken into account separately. These residual skip connections are those where the output of an arbitrary layer $s$ is added to the output of another layer $h$ connected to the following layer ${h+1}$, where the residual connections converge. To account for these, the Ratio-Based splitting proposed in \cite{otsuki_layer-wise_2024} is adopted, whereby the relevance is distributed according to the ratio between $\sigma^{(s)}$ and $\sigma^{(h)}$, which indicate the activation of the skip connection from layer $s$ and the activation of layer $h$, respectively. Following this approach, the contribution of an arbitrary neuron $i$ in layer ${h+1}$, $R^{({h+1})}_i$, is divided as follows:
\begin{equation}
    R^{(s)}_i= \frac{R^{({h+1})}_i\mid \sigma^{(s)}_i \mid}{\mid \sigma^{(s)}_i \mid + \mid \sigma^{(h)}_i \mid + \epsilon}, 
    \quad 
    R^{(h)}_i= \frac{R^{({h+1})}_i\mid \sigma^{(h)}_i \mid}{\mid \sigma^{(s)}_i \mid + \mid \sigma^{(h)}_i\mid + \epsilon}
\end{equation}
where $\epsilon$ is also taken into account, in contrast to the proposal of \cite{otsuki_layer-wise_2024}. This adjustment offers numerical stability when activations are negligible or very close to zero, which constitutes a significant challenge when using the DaFFs, as they are designed to ensure that all activations are set to zero across all layers at the boundary condition collocation points. Therefore, when the backpropagation reaches $R^(s)$, the addition of $R^{(h)}_i$ and $R^{(h+1)}_i$ is sufficient in order to ensure the conservation property. 

In this work, there is a differentiation between interpretability and explainability, whereas interpretability is related to the intrinsic structure of the model or its input representation that allows human-understandable insight into how the model functions, while explainability refers to the extraction of human-understandable explanations from the trained models, usually via post-hoc methods that analyze the models. Note that, although both are connected as more interpretable models ease the extraction and construction of explanations, they are not equivalent, and interpretability alone does not guarantee that explanations can be extracted without dedicated post-hoc techniques. In our setting, DaFFs provide a more interpretable spectral representation than RFFs due to their physically grounded structure, while LRP serves as the explainability mechanism used to extract relevance information from the trained models.

\section{Numerical examples}
\label{sec:numerical-examples}

In this section, the performance of the PINN-DaFFs model is investigated by means of two numerical examples: the Kirchhoff-Love and the Helmholtz equations. Although \cref{sec:problem-description} presents the general space-time formulation, both benchmark problems considered here are stationary and depend only on the spatial coordinates. We compare the performance and accuracy of the proposed model, PINN-DaFFs, against vanilla PINNs and PINN-RFFs. By design, the training of these models follows a zero-data approach, since the optimization is performed using PDE-related residuals, which are computed for collocation points within the domain, as presented in \cref{sec:problem-description}. The number of collocation points denotes the batch size, meaning that higher batch sizes enrich the training with more data in each epoch, followed by a consequent increase in the computational cost. In the analysis results presented below, 3/4 of the collocation points are distributed in the interior of the domain $\Omega$, while the remaining $1/4$ is distributed in the boundary $\partial\Omega$.

The hyperparameter tuning process was carried out in two phases. A manual exploratory step was first adopted for the selection of an initial configuration, and subsequently, a random search was used for the exploration of the search space, where an increased range of values was centered around the initial configuration. The number of training trials is hereafter referred to as \textit{random samples}, together with the total number of possible combinations. The hyperparameters considered are the layers, units per layer, batch size, and learning rate, which have been optimized for all the compared PINNs variants.

For PINN-RFFs, the variance of the Fourier features is a model-specific hyperparameter indicated as \textit{Fourier $\sigma^2_j$} $= (\sigma^2_1,\sigma^2_2,\dots,\sigma^2_d)$. The \textit{Fourier $\sigma^2_j$} hyperparameter encodes two things. First, the number of values $d$ indicates the number of RFFs layers concatenated, and second, each numerical value denotes the variance $\sigma^2_j$ from which to sample $\mathbf{b}_j$, as shown in \cref{eq:random_FF}. This means, in terms of input dimensionality, that PINN-RFFs have a dimensionality in their RFF layer equal to $2\times d$ times the number of hidden units, denoted by the \textit{Units} hyperparameter.

For PINN-DaFFs, the model-specific \textit{DaFF comp} hyperparameter determines the kind of signals combined to form the DaFF, while the $(m,n)$ parameters from \textit{DaFF $(m,n)$} control the frequency. These model-specific hyperparameters are depicted as sets of values within $(\cdot)$. \textit{DaFF comp} represents the number of DaFF layers concatenated, and each value is the type of component, while \textit{DaFF $(m,n)$} refers to the pair of $(m,n)$ parameters from which to make the permutations that are used to define the DaFFs, where $m \in \{1, 2,  \ldots, m_N\}$ and $n \in \{1, 2,  \ldots, n_N\}$. It should be noted that the listed model-specific hyperparameter values for PINN-RFFs and PINN-DaFFs were chosen to span progressively richer and more complex feature representations while keeping the random-search space computationally manageable, rather than to account for all intermediate combinations.

For the models that contain multiple loss terms, an adaptive loss balancing strategy was adopted, namely the Relative Loss Balancing with Random Lookback (ReLoBRaLo), to deal with gradient stiffness issues associated with vanilla PINNs. Within this context, the different loss terms related to the PDE and the boundary conditions are dynamically adapted during the training phase. All models were trained for $50000$ epochs, consisting of 5000 steps of Broyden–Fletcher–Goldfarb–Shanno (BFGS) algorithm. A learning rate decay strategy was used, with a factor of $0.1$, after a $2000$ epochs patience period, and an early stopping with patience equal to $2000\times 2+1$. 

The reported training times correspond to the neural-network training process, and therefore include the repeated evaluation of the selected input representation. In the case of PINN-DaFFs, this means that the cost of mapping the inputs through the DaFF transformation is included in the reported values, just as the corresponding RFF mapping is included for PINN-RFFs. However, the one-time construction of the DaFF basis itself, obtained from the Laplace eigenvalue problem of the domain either analytically or numerically, is performed offline before training and is therefore not included in these times. Consequently, the additional computational overhead of the proposed approach lies in this offline DaFF construction step, while the training-time differences reported below reflect the cost of optimizing the resulting models with their respective input representations.

\subsection{Kirchhoff-Love Equation}
\label{subsec:kirchhoff}

The bending of plate structures, referring to the deflection of plates perpendicular to their plane due to the action of external loads, is a widely confronted problem in structural and mechanical systems and is described by the Kirchhoff-Love theory of plates \cite{kirchhoff_uber_1850, love_xvi_1997}. Within this context, three-dimensional plates are represented by a mid-surface plane, whose out-of-plane displacement field is described by the following fourth-order PDE 

\begin{subequations}
    \begin{align}
        \nabla^4 u(x, y) = \frac{f(x, y)}{D}, & \quad 
        (x, y)\in\Omega\label{eq:kirchhoff-pde}\\[3mm]
        u(x, y) = 0, & \quad
        (x, y) \in \partial\Omega\label{eq:kirchhoff-bc1}\\[3mm]
        -D\left(\frac{\partial^2 u}{\partial x^2} +
        \nu \frac{\partial^2 u}{\partial y^2}\right) = 0, & \quad
        (x, y) \in \partial\Omega_1 \cup \partial\Omega_3\label{eq:kirchhoff-bc2}\\[2mm]
        -D\left(\nu\frac{\partial^2 u}{\partial x^2} +
        \frac{\partial^2 u}{\partial y^2}\right) = 0, & \quad
        (x, y) \in \partial\Omega_2 \cup \partial\Omega_4\label{eq:kirchhoff-bc3}
    \end{align}
    \label{eq:kirchhoff}
\end{subequations}

defined in a rectangular domain $\Omega=[0, a]\times [0, b]$, with $\partial \Omega_1$, $\partial\Omega_3$ denoting the boundaries at $x=0$ and $x=a$ respectively, while $\partial \Omega_2$, $\partial\Omega_4$ are the boundaries at $y=0$ and $y=b$ respectively. In \cref{eq:kirchhoff}, $D$ is the flexural stiffness of the plate, depending on the material Young's modulus $E$, the thickness of the domain $h$, and Poisson's ratio $\nu$. Lastly, $f(x, y)$ denotes the distributed load acting on the entire domain $\Omega$. These parameters are given by the following expressions
\begin{subequations}
    \begin{align}
        D &= \frac{Eh^3}{12(1-\nu)}\\[2mm]
        f(x,y) &= f_0 
        \sin\left(\frac{x\pi}{a}\right)
        \sin\left(\frac{y\pi}{b}\right)
    \end{align}
\end{subequations}
where $f_0$ indicates the magnitude of the external load, while $a$ and $b$ are the dimensions of the plate in $x$ and $y$ directions respectively. Notably, these boundary conditions include both function values and derivative terms. Thus, the examples implicitly demonstrate that both Dirichlet and Neumann boundary conditions can be inherently supported by the proposed DaFFs.

\begin{table}[!t]
    \footnotesize
    \centering
    \caption{Hyperparameter and best performing model summary for the Kirchhoff-Love problem. Configurations with the best performance for each model are marked in bold. The scale disparity between training and validation losses is due to the training loss being the average loss between PDE and boundary condition losses, while the validation loss is the MSE with respect to the solution at each collocation point. Loss metrics for the best-performing model are displayed first, followed by the mean losses and standard deviations (STD) across 10 models trained with the best configuration.}
    \label{tab:hyperparameter_search}
    \begin{tabular}{l V{4} c|c|c}
         & PINN & PINN-RFFs & PINN-DaFFs \\
        \hlineB{4}
        Layers & 2,\textbf{3},4,5 & \textbf{2},3,4,5 & 2,\textbf{3},4\\
        Units & 64, \textbf{128}, 256 & 64, \textbf{128}, 256 & 64, 128, \textbf{256}\\
        Batch size & 512, 1024, \textbf{2048} & 512, 1024, \textbf{2048} & 512, 1024, \textbf{2048}\\
        Learning rate & 0.005, 0.001, \textbf{0.0005}, 0.0001 & 0.005, 0.001, 0.0005, \textbf{0.0001} & 0.005, \textbf{0.001}, 0.0005\\
        \textit{RFF $\sigma^2_j$} & N/A & \textbf{(1)}, (1,2), (1,2,3,4) & N/A\\
        \textit{DaFF comp} & N/A & N/A & (1,3), (2,4), \textbf{(1,2,3,4)} \\
        \textit{DaFF $(m,n)$} & N/A & N/A & (1), \textbf{(1,-1)}, (1,2,3,4,5) \\
        Loss Balancing & Yes & Yes & No \\
        Bias term & Yes & Yes & No \\
        Random samples & 30 of 144 & 40 of 432 & 20 of 729 \\
        \hline
        Best train loss & $2.71\times10^{-04}$ & $7.74\times10^{-06}$ & $1.02\times10^{-09}$ \\
        Best val loss & $2.75\times10^{-06}$ & $3.01\times10^{-07}$ & $6.42\times10^{-18}$ \\
        Best train time & 01h 03m 04s & 44m 09s & 45m 51s \\
        
        \hline
        Mean $\pm$ STD train loss & 
        {\fontsize{6}{7.2} $1.11\times10^{-03}\pm4.69\times10^{-04}$} & 
        {\fontsize{6}{7.2} $1.49\times10^{-05}\pm7.33\times10^{-06}$} & 
        {\fontsize{6}{7.2} $3.68\times10^{-08}\pm4.70\times10^{-08}$} \\
        
        Mean $\pm$ STD val loss & 
        {\fontsize{6}{7.2} $2.66\times10^{-05}\pm3.59\times10^{-05}$} & 
        {\fontsize{6}{7.2} $1.69\times10^{-06}\pm1.82\times10^{-06}$} & 
        {\fontsize{6}{7.2} $1.20\times10^{-17}\pm1.34\times10^{-17}$} \\
    \end{tabular}
\end{table}

\begin{figure}[!htbp]
    \centering
    \includegraphics[width=0.95\textwidth]{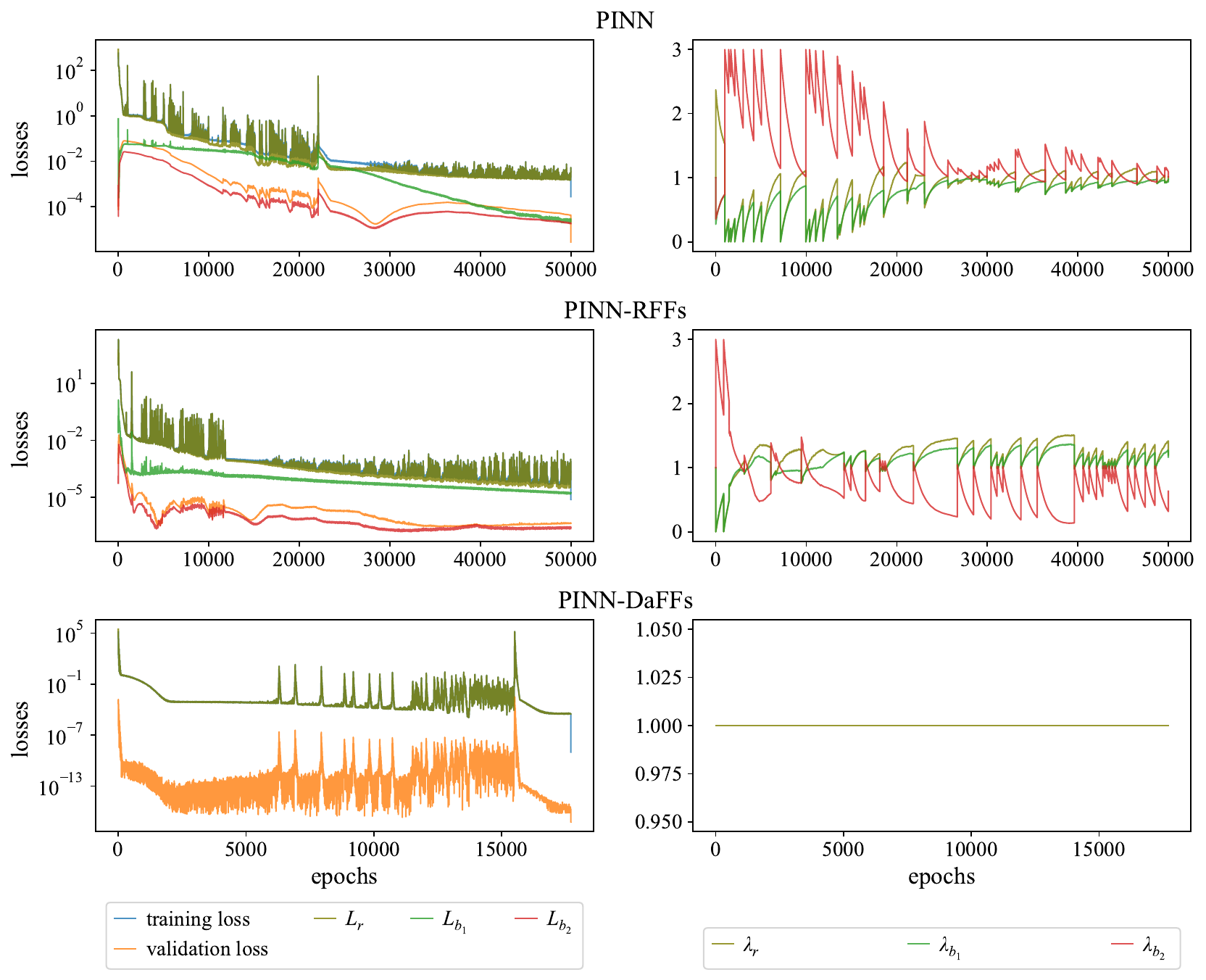}
    \caption{Learning phase for the Kirchhoff problem; the left column depicts the evolution of the training and validation losses over epochs; the PDE loss denoted by $L_{\mathrm{r}}$ and the boundary condition loss terms denoted by $L_{\mathrm{b}_1}$ and $L_{\mathrm{b}_2}$; the right column presents the weights assigned to the loss terms by the loss balancing strategy during training.}
    \label{fig:training_kirch}
\end{figure}

Given the above problem definition, the loss function is decomposed into the following components
\begin{equation}
    L = \lambda_{\mathrm{r}}L_{\mathrm{r}}(\bm{\theta}) + 
    \lambda_{\mathrm{b}_1}L_{\mathrm{b}_1}(\bm{\theta}) + 
    \lambda_{\mathrm{b}_2}L_{\mathrm{b}_2}(\bm{\theta})
\end{equation}
where $L_{\mathrm{r}}$ denotes the loss term associated with \cref{eq:kirchhoff-pde}, while the boundary condition loss terms are divided into two parts, $L_{b_1}$ associated with \cref{eq:kirchhoff-bc2,eq:kirchhoff-bc3}, and $L_{b_2}$ related to \cref{eq:kirchhoff-bc1}. It should be underlined that the introduced DaFFs do not require the minimization of the boundary condition terms, as these are satisfied by definition. This implies the loss function for the PINN-DaFFs model, which contains only the $L_{\mathrm{r}}$ term, thus simplifying the training phase without requiring the use of a loss balancing scheme.

A summary of the explored hyperparameters, related to performance, and training time is presented in \cref{tab:hyperparameter_search}, with the values corresponding to the best configuration shown in bold fonts. The proposed PINN-DaFFs model is several orders of magnitude better in performance than the other PINN variants, with similar or less computation time required for training, and significantly fewer training epochs required for effective learning. This performance improvement is observed not only in terms of the training error, which is computed by the aggregation of the PDEs and boundary condition loss terms (boundary condition losses only required for vanilla PINN and PINN-RFFs), but also in terms of the validation loss, computed as the MSE between predicted and true value for a grid of points within the domain of the problem. The scale difference between validation and training errors is due to the fact that the former is the MSE between the models' outputs and ground truth, which is essentially the analytical solution of the PDE, while the latter is calculated from the PDE residuals that are dependent on the physics equations, namely \cref{eq:loss}, and do not include the ground truth. This applies to both numerical examples examined in this work. Moreover, the mean training and validation losses, and their standard deviations, computed over 10 runs of the best configuration, indicate that the observed performance improvements are consistent in this experiment, with PINN-DaFFs showing lower variability across runs, suggesting more robustness to training stochasticity.

The training phase corresponding to the best configuration for each model variant is presented in \cref{fig:training_kirch}, with the learning curves shown on the left side, and the weights of the loss balancing scheme on the right side. In all three cases, the PDE loss term, $L_r$, displays an unstable behavior, while the other loss components, which are present only for the PINN and PINN-RFFs models, are much more stable. It is further shown that the loss balancing strategy plays a significant role in the training, where the boundary condition terms are much stiffer, hence requiring higher weight values. The inclusion of both types of Fourier features allows the models to achieve good results quickly for some of the loss terms. However, the existence of multiple loss terms complicates the training process for the PINN-RFFs model, which requires the addition of a loss lancing scheme. On the other hand, the PINN-DaFFs training requires only the PDE loss term, and although the learning curve may appear to be characterized by higher instability, the scale of the losses is significantly smaller, thus resulting in much faster learning.

\subsection{Helmholtz Equation}
\label{subsec:helmholtz}

The Helmholtz equation, also known as the Poisson-Boltzmann equation, describes a time-independent form of the wave equation, relevant to many physics problems. The equation is herein defined in two dimensions, in the domain $\Omega=[-1, 1]^2$, and has the following form
\begin{subequations}
    \begin{align}
        \Delta u(x, y) + k^2 u(x, y) = f(x, y),& \quad (x, y) \in \Omega
        \label{eq:helmholtz}\\[2mm]
        u(x, y) = 0,& \quad (x, y) \in \partial\Omega
    \end{align}
\end{subequations}
where

\begin{equation}
    f(x, y) = (-(n_1\pi)^2 - (n_2\pi)^2 + k^2)\sin(n_1 \pi x)\sin(n_2 \pi x)
\end{equation}

with $n_1$ and $n_2$ denoting the input harmonic numbers in $x$ and $y$ directions respectively. For this problem, the boundary conditions are split into four equations
\begin{subequations}
    \begin{align}
        u(x, y) & = 0, \quad (x, y) \in \partial\Omega_1 = [-1, 1] \times \{-1\}\\[1mm]
        u(x, y) & = 0, \quad (x, y) \in \partial\Omega_2 = [-1, 1] \times \{1\}\\[1mm]
        u(x, y) & = 0, \quad (x, y) \in \partial\Omega_3 = \{-1\}\times [-1, 1]\\[1mm]
        u(x, y) & = 0, \quad (x, y) \in \partial\Omega_4 = \{1\}\times [-1, 1]
    \end{align}
    \label{eq:helmholtz-individual-bcs}
\end{subequations}

referring to each one of the domain edges. According to this separation, the loss function is accordingly decomposed in the following terms

\begin{equation}
    L = \lambda_{\mathrm{r}}L_{\mathrm{r}}(\bm{\theta}) + 
    \lambda_{\mathrm{b}_1}L_{\mathrm{b}_1}(\bm{\theta}) + 
    \lambda_{\mathrm{b}_2}L_{\mathrm{b}_2}(\bm{\theta}) +
    \lambda_{\mathrm{b}_3}L_{\mathrm{b}_3}(\bm{\theta}) + 
    \lambda_{\mathrm{b}_4}L_{\mathrm{b}_4}(\bm{\theta})
\end{equation}

where $L_{\mathrm{r}}$ is the PDE loss, derived from \cref{eq:helmholtz}, and $L_{\mathrm{b}_j}$, for $j=1,2,3,4$, are the loss terms associated with \cref{eq:helmholtz-individual-bcs}. Again, it should be noted that these terms are omitted in the training of the PINN-DaFFs model and are relevant only for the other two variants.


\begin{table}[!htbp]
    \footnotesize
    \centering
    \caption{Hyperparameter and best performing model summary for the Helmholtz problem. Configurations with the best performance for each model are shown in bold. The scale disparity between training and validation losses is due to the training loss being the average loss between PDE and boundary condition losses, while the validation loss is the MSE with respect to the solution at each collocation point. Loss metrics for the best-performing model are displayed first, followed by the mean losses and standard deviations across 10 models trained with the best configuration.}
    \label{tab:hyperparameter_search_helm}
    \begin{tabular}{l V{4} c|c|c}
         & PINN & PINN-RFFs & PINN-DaFFs \\
        \hlineB{4}
        Layers & 2,3,4,\textbf{5} & 2,\textbf{3},4,5 & 3,4,\textbf{5} \\
        Units & 64, 128, \textbf{256} & 64, \textbf{128}, 256 & 64, \textbf{128}, 256\\
        Batch size & 512, \textbf{1024}, 2048 & 512, 1024, \textbf{2048} & 512, \textbf{1024}, 2048\\
        Learning rate & \textbf{0.005}, 0.001, 0.0005, 0.0001 & 0.005, 0.001, \textbf{0.0005} & 0.001, \textbf{0.0005}\\
        \textit{Fourier $\sigma^2_j$} & N/A & \textbf{(1)}, (1,2,3), (1,2,3,4,5) & N/A\\
        \textit{DaFF comp} & N/A & N/A & (1), \textbf{(1,2)} \\
        \textit{DaFF $(m,n)$} & N/A & N/A & \begin{tabular}[t]{@{}c@{}c@{}c@{}}(1), \textbf{(2,8)}, (2,8,16,32,64), \\ (1,2,8,16,32,64,128,256), \\ (1,2,3,4,5,6,7,8,9,10 \\ ,11,12,13,14,15) \end{tabular} \\
        Loss Balancing & Yes & Yes & No \\
        Bias term & Yes & Yes & No \\
        Random samples & 30 of 144 & 35 of 324 & 40 of 540 \\
        \hline
        Best train loss & $1.23\times10^{-03}$ & $8.83\times10^{-04}$ & $3.86\times10^{-05}$ \\
        Best val loss & $5.48\times10^{-05}$ & $7.75\times10^{-06}$ & $2.32\times10^{-11}$ \\
        Best train time & 02h 51m 25s & 01h 13m 06s & 31m 25s \\

        \hline
        Mean $\pm$ STD train loss & 
        {\fontsize{6}{7.2} $6.65\times10^{-01}\pm1.64\times10^{0}$} & 
        {\fontsize{6}{7.2} $1.08\times10^{-03}\pm3.21\times10^{-04}$} & 
        {\fontsize{6}{7.2} $3.09\times10^{-04}\pm3.01\times10^{-04}$} \\
        
        Mean $\pm$ STD val loss & 
        {\fontsize{6}{7.2} $1.02\times10^{-03}\pm1.63\times10^{-03}$} & 
        {\fontsize{6}{7.2} $1.50\times10^{-05}\pm4.47\times10^{-06}$} & 
        {\fontsize{6}{7.2} $1.49\times10^{-09}\pm2.63\times10^{-09}$} \\
    \end{tabular}
\end{table}

The hyperparameter tuning process for the Helmholtz problem is analogous to the one followed in the previous case study for the Kirchhoff problem. The hyperparameter combinations tested through the random search phase are presented in \cref{tab:hyperparameter_search_helm} along with the best configurations for each model, shown in bold face, and the corresponding loss values and training times. Both models relying on Fourier features result in a significant reduction of the training time, which becomes significantly lower when using DaFFs. Furthermore, the performance of PINN-DaFFs demonstrates a substantial improvement in both training and validation losses, which are a couple of orders of magnitude smaller than those achieved by the other two models. Regarding mean and standard deviation metrics, likewise computed over 10 runs of the best configuration, a similar conclusion can be obtained, as the statistics confirm the performance advantage of PINN-DaFFs, together with lower variability across runs than in the other models.

\begin{figure}[!htbp]
    \centering
    \includegraphics[width=0.95\textwidth]{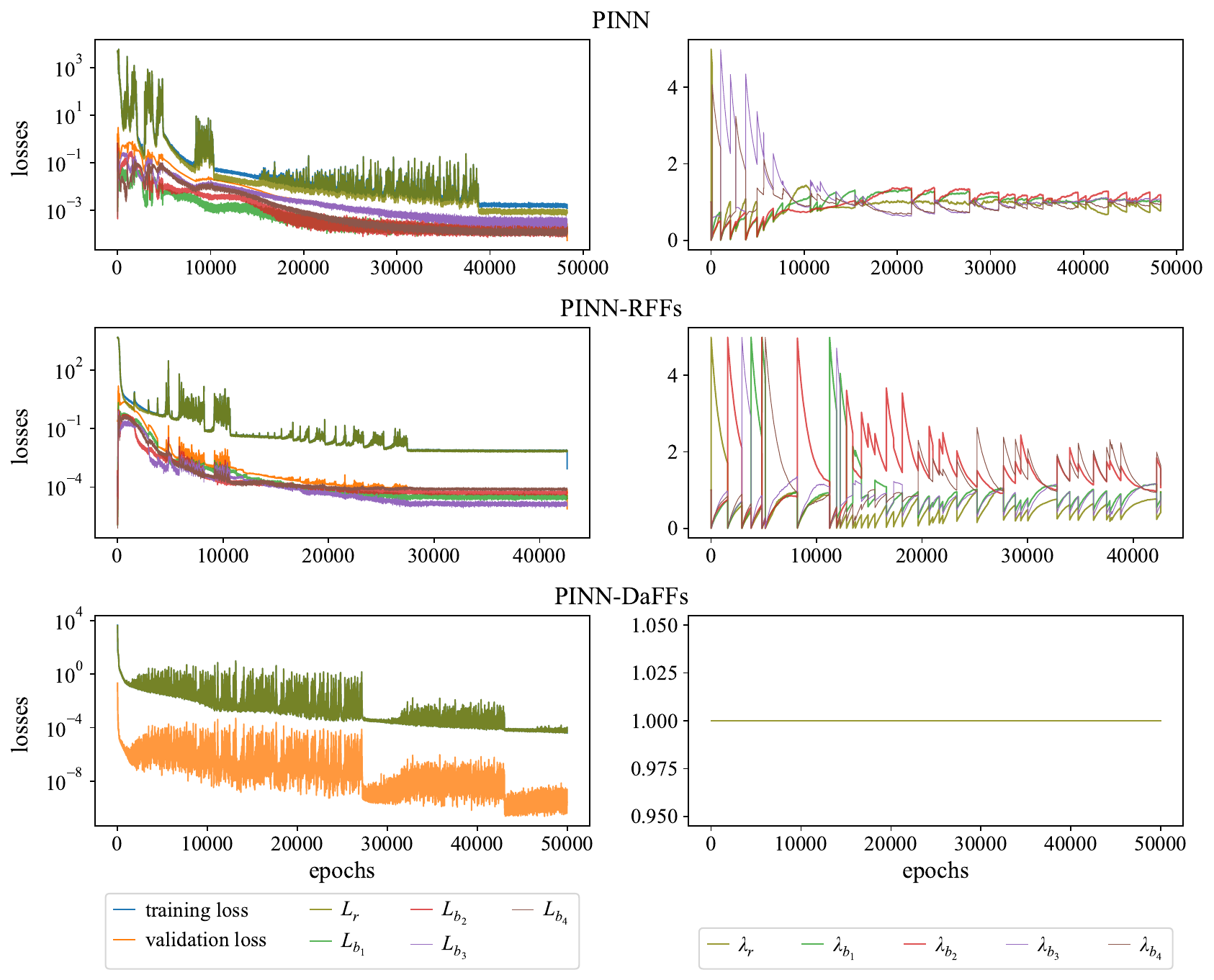}
    \caption{Learning phase for the Helmholtz problem; the left column depicts the evolution of the training and validation losses over epochs; the PDE loss is denoted by $L_{\mathrm{r}}$ and the loss terms of the boundary condition loss terms are denoted by $L_{\mathrm{b}_1}$, $L_{\mathrm{b}_2}$, $L_{\mathrm{b}_3}$ and $L_{\mathrm{b}_4}$; the right column presents the weights assigned to the loss terms by the loss balancing strategy during training}
    \label{fig:training_helm}
\end{figure}

The learning curve of the best configuration for each one of the model variants is presented in \cref{fig:training_helm}. The vanilla PINN model shows how the large number of loss terms poses a significant challenge during early training phases, even with the help of the loss balancing scheme. Although to a lesser extent, the same pattern is observed with the PINN-RFFs model. On the contrary, the PINN-DaFFs model relies on a single objective and, as such, does not suffer from this particular issue. Regarding the model loss, it decreases relatively unstably with vanilla PINNs, which seems to be improved with both types of Fourier features (FFs), although DaFFs show better results considering the scales. Although the early stopping criterion is satisfied only for the PINN-RFFs model, the required training time is double compared to the one required by the PINN-DaFFs model, which is due to the higher input dimensionality and the higher batch size (more data used for training) used for PINN-RFFs. 

\subsection{XAI Analysis}
\label{subsec:xai}

\begin{figure}[!htbp]
    \centering
    \subfloat[Kirchhoff.]
    {\includegraphics[width=0.4\textwidth]{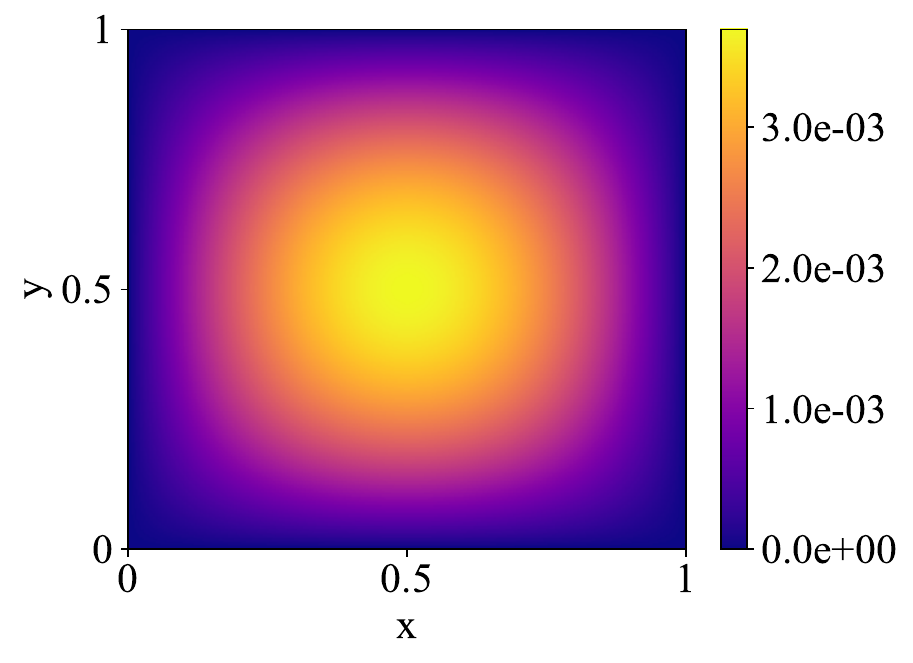}\label{fig:kirchhoff_solution}}
    \centering
    \subfloat[Helmholtz.]
    {\includegraphics[width=0.4\textwidth]{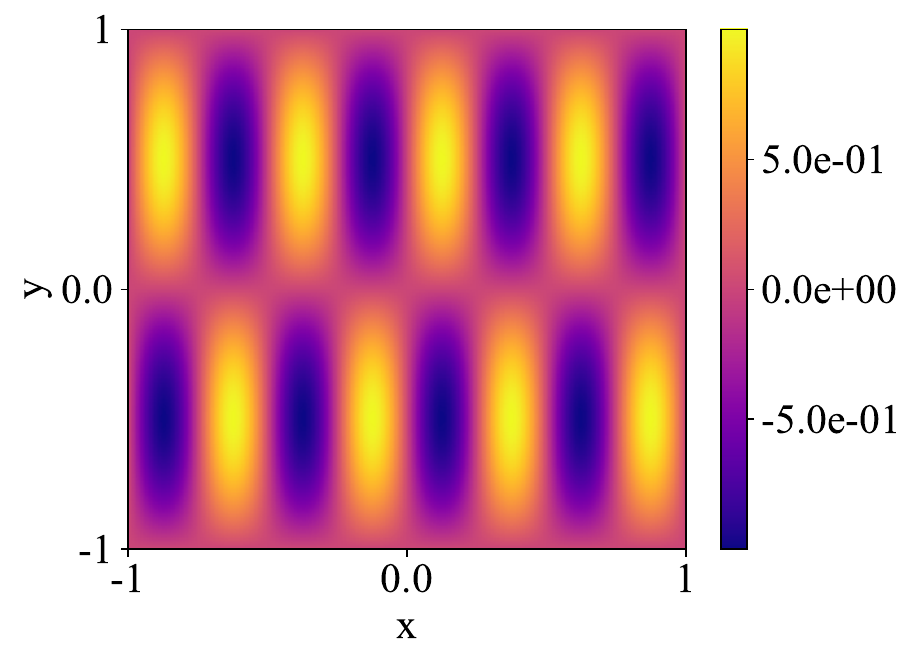}\label{fig:helmholtz_solution}}
    \caption{Solution to the Kirchhoff and Helmholtz numerical examples. Given the value distribution, the spectral components for Kirchhoff should approximate $sin(b*x)$ and $ cos(b*x)$ where $x \in \Omega, b \approx \pi/2$, while for Helmholtz they are dependent on the coordinate, with $b \approx \pi$ for the $y$ component and $b \approx 4\pi$ for the $x$ component.}
    \label{fig:solutions}
\end{figure}

\begin{figure*}[!htbp]
    \centering
    \subfloat[LRP of best vanilla PINN model for Kirchhoff.]
        {\includegraphics[width=0.9\linewidth]{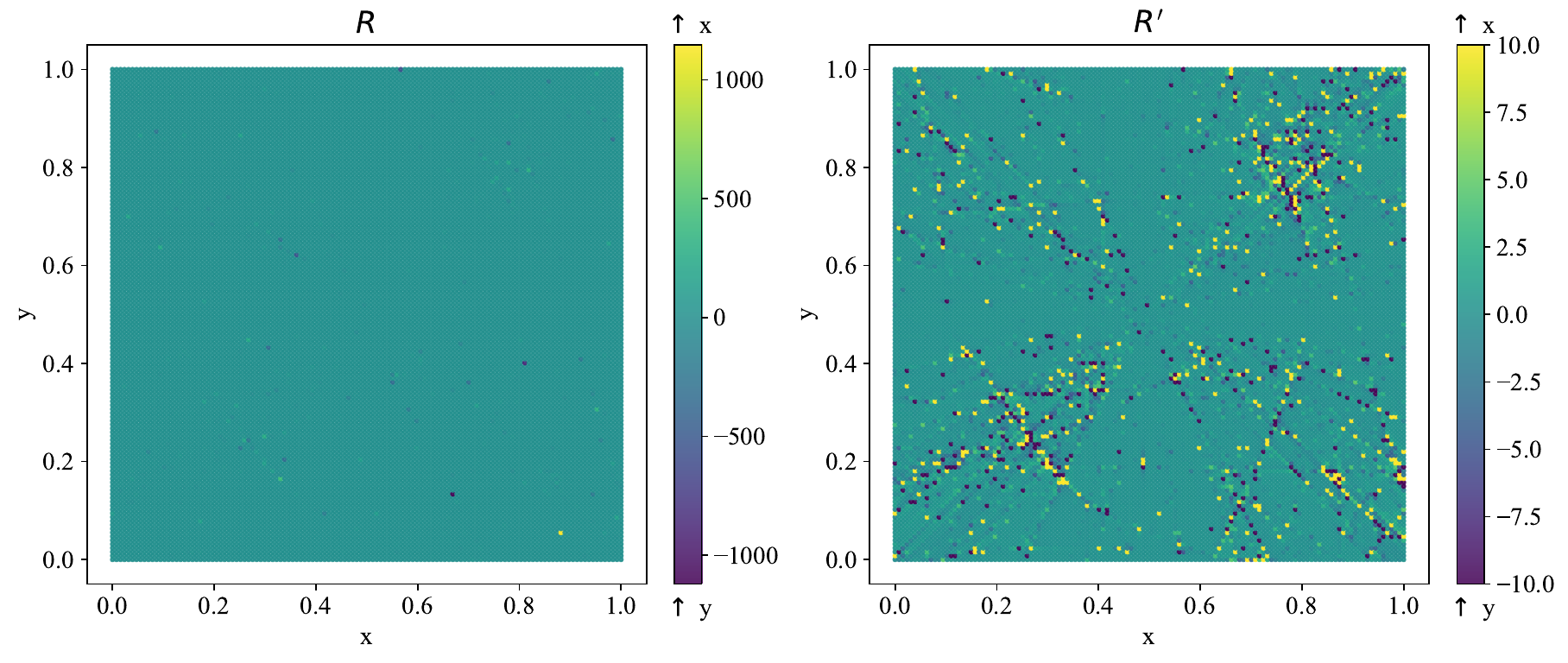}\label{fig:lrp_kirch_PINN_filtered}}\\
    \centering
    \subfloat[LRP of best vanilla PINN model for Helmholtz.]
        {\includegraphics[width=0.9\linewidth]{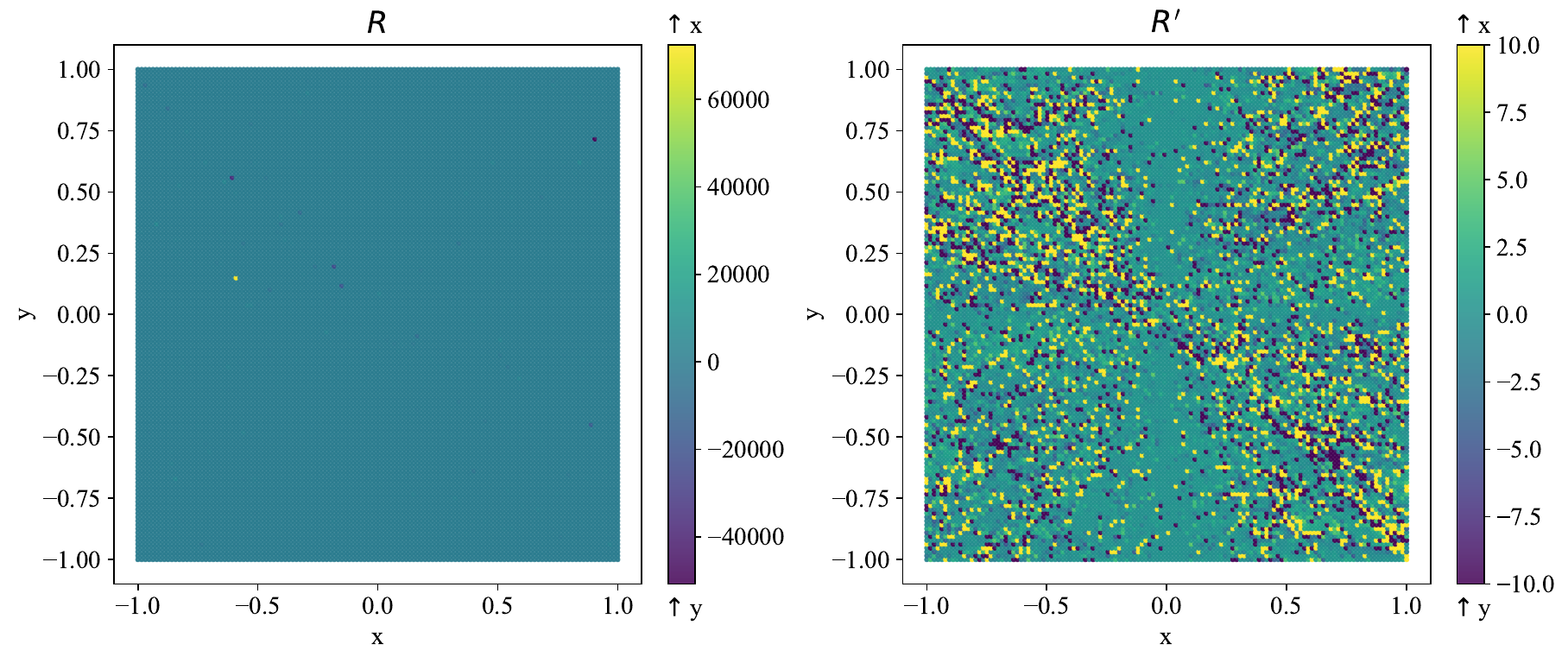}\label{fig:lrp_helm_PINN_filtered}}\\
    \caption{Contribution of the $x$ and $y$ inputs to the prediction of the best vanilla PINN across the whole domain of the problems computed with LRP. On the left, $R$ is computed by LRP, where positive values indicate a stronger effect from $x$ and negative values indicate a higher impact from $y$. On the right, the contributions $R'$ of the PINN are filtered with an arbitrary threshold of 10, so that $if \quad |R(x,y)| > 10, \quad then |R(x,y)| = 10, \quad \forall (x,y) \in \Omega$.}
    \label{fig:lrp_PINN_filtered}
\end{figure*}

For the assessment of the analysis results and the performance of the proposed model in comparison to the vanilla PINN and PINN-RFFs models, the LRP method is used in order to gain further insights into the contribution of each input to the final prediction. For a vanilla PINN model, we can examine the predictions for the Kirchhoff and Helmholtz problems in a grid within the problem's domain. From \cref{eq:kirchhoff,eq:helmholtz}, we expect that the spatial coordinates have similar contributions to the solution since they are symmetrical problems, as shown in \cref{fig:solutions}, where the target solution of each problem is displayed. The contribution scores of the input coordinates, $x$ and $y$, are presented in \cref{fig:lrp_PINN_filtered} for the best configuration of the vanilla PINN model.

In \cref{fig:lrp_PINN_filtered}, it is shown that there are many points in each of the problems' domains with significant differences between each coordinate's contribution. This pattern is more clearly evidenced when a threshold of $10$ is applied to the contribution colormap, which can be an indication of the poor expressiveness and interpretability offered by vanilla PINNs. Moreover, there are some sparse points where the differences are extreme, which can be observed in the left-side plots of \cref{fig:lrp_PINN_filtered}. These results can be helpful to elucidate flaws in the model's predictions of the simulated dynamics and to understand where these should be treated more carefully due to uncertainties in the compliance with the known physics, despite performance metrics. However, it is also clear that the potential for interpretability of models with inputs that consist only of spatial coordinates is rather poor, and reaching clear conclusions is a challenging task. 

It should be underlined that with the use of Fourier features, the dimensionality of the input space increases due to the encoding of the input coordinates with a larger number of random harmonic features. This spatial coordinate transformation of the collocation points is carried by a non-injective function which, depending on the sampled $b_j$ parameter values of \cref{eq:random_FF}, may deliver the same result for different $\mathbf{x}$ values. Therefore, for the PINN-RFFs and PINN-DaFFs, the focus is on understanding the relevance and contribution of each Fourier feature in the learning process, with the aim of unraveling how these are used by the model. Moreover, the use of Fourier features requires the optimization of some hyperparameters, such as the variance $\sigma^2_{j,k}$ of each feature coefficient $\mathbf{b}_{j,k}$ for RFFs, and the set of harmonic indices $(m_j,n_j)$ for each $\phi_j(\mathbf{x})$ of the DaFFs. 

\begin{figure*}[!htb]
    \centering
    \subfloat[Best PINN-RFFs model configuration.]
        {\includegraphics[trim={0 1.3cm 0 0},clip,width=0.95\linewidth]{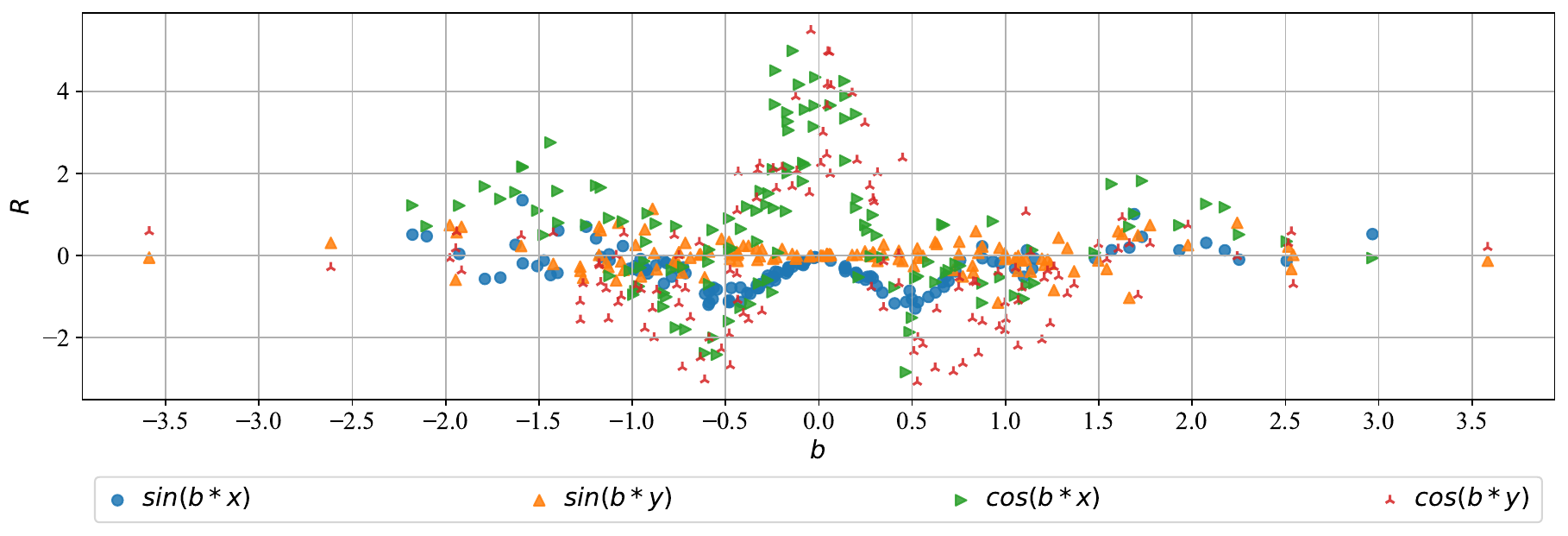}\label{fig:lrp_kirch_RFF_PINN_scatter}}\\
    \subfloat[PINN-RFFs model with extended \textit{Fourier $\sigma^2_j$} = (1,2,3,4,5).]
        {\includegraphics[trim={0 1.3cm 0 0},clip,width=0.95\linewidth]{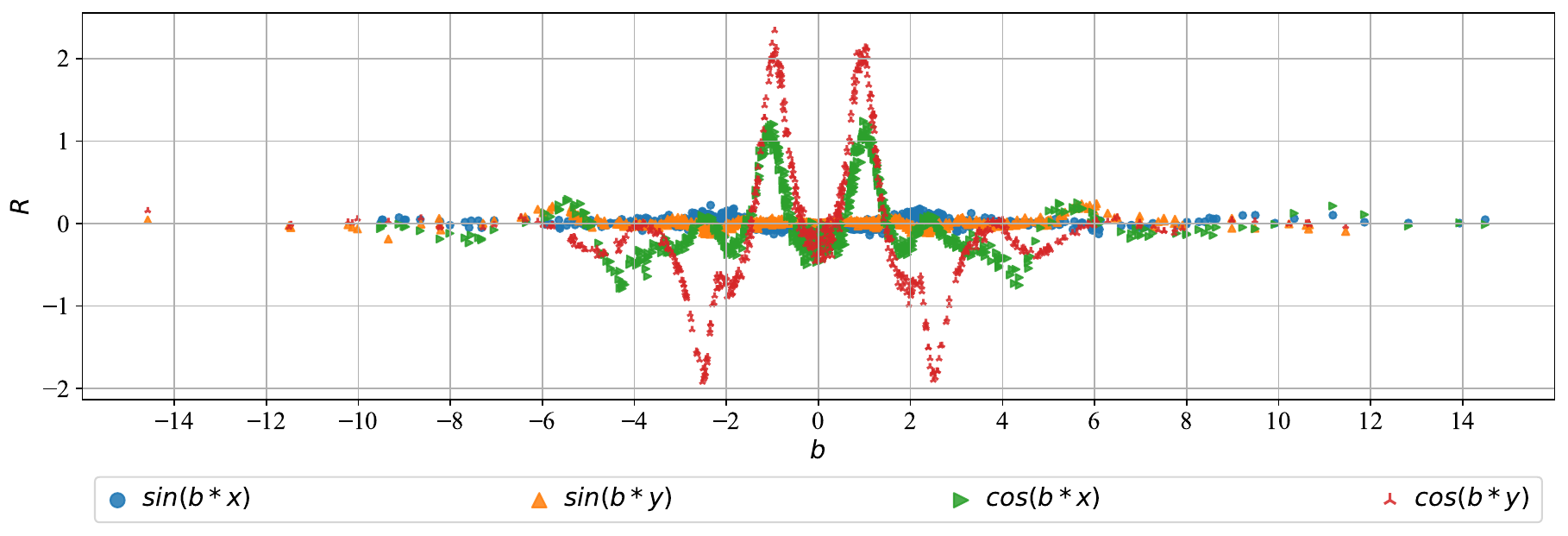}\label{fig:lrp_kirch_RFF_PINN_scatter_fs_1}}\\
    \centering
    \subfloat[PINN-RFFs model with extended \textit{Fourier $\sigma^2_j$} = (1,5,10).]
        {\includegraphics[width=0.95\linewidth]{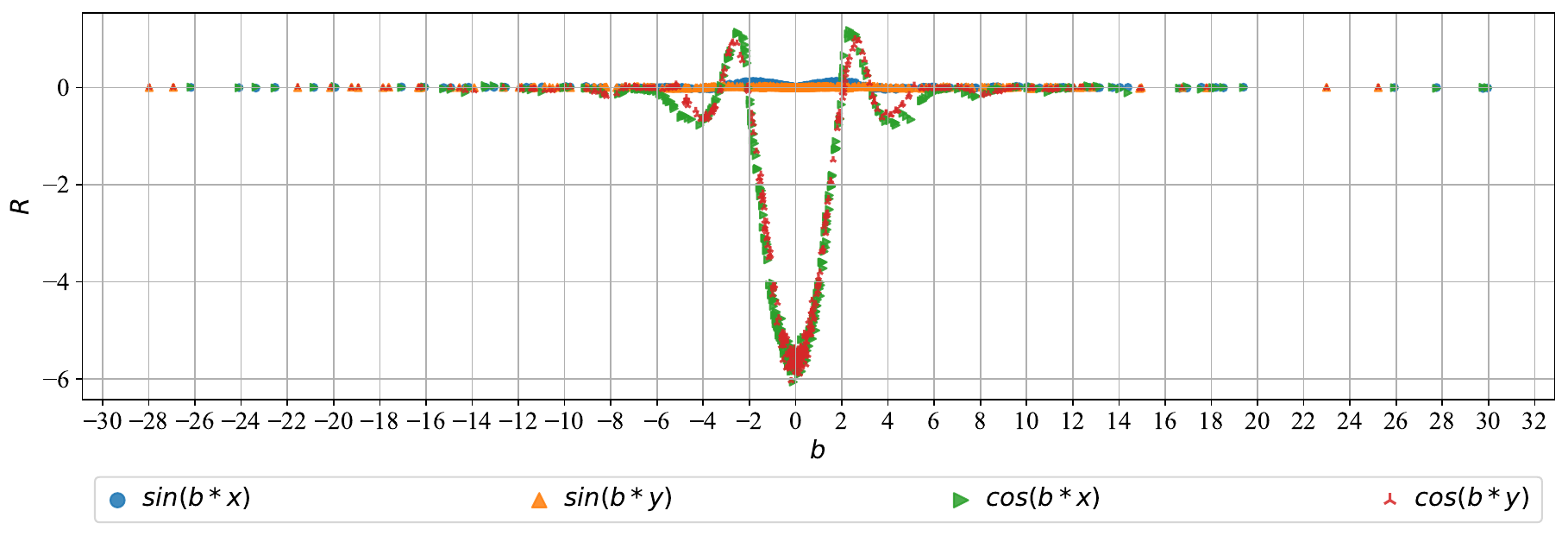}\label{fig:lrp_kirch_RFF_PINN_scatter_fs_2}}\\
    \caption{Kirchhoff-love case: contribution $R$ of RFFs to the predictions of PINNs with respect to the $\mathbf{b}$ parameters and the type of Fourier component and spatial coordinate. No clear correlations can be found between models with different RFF sampling parameters, suggesting a shortage of generalizable and learnable patterns.}
    \label{fig:PINNs_RFF_kirch}
\end{figure*}

Under this perspective, it is shown herein that the LRP analysis can be used as a means of guidance for feature selection. For this purpose, the contribution $R$ of the RFFs from three different PINN-RFFs models is presented in \cref{fig:PINNs_RFF_kirch}, in which the $x$-axis represents the $\mathbf{b}$ value corresponding to each RFF. It is observed that there are three main regions, or peaks, of the RFFs that carry the highest contribution of the best-performing model, whose RFF contributions are shown in \cref{fig:lrp_kirch_RFF_PINN_scatter}. Sorted from higher to lower, the first peak is centered around $[-0.25,0.25]$, the second one around $[-0.8,-0.5] \cup [0.5,0.8]$, and the third one around $[-1.75,-1.25] \cup [1.5,1.75]$. Comparing the type of features, the cosine-based holds most of the contribution for both $x$ and $y$, with an average over the absolute value of the contributions of $1.61$ and $1.82$, respectively. On the other hand, the sin-based features offer a contribution of around $0.46$ and $0.35$ for $x$ and $y$, respectively.

The other two examples in \cref{fig:lrp_kirch_RFF_PINN_scatter_fs_1}, where 5 RFFs are used with variance $\sigma^2_j=(1,2,3,4,5)$, and \cref{fig:lrp_kirch_RFF_PINN_scatter_fs_2}, where three RFFs with $\sigma^2_j=(1,5,10)$ are used, cover a larger range of the $\mathbf{b}$ space with more sample points, each of which corresponds to a different Fourier feature. In both of them, the cosine signals hold most of the contribution, but the contribution peaks are located around different values. In \cref{fig:lrp_kirch_RFF_PINN_scatter_fs_1}, there are two high peaks around $b\approx\pm1$, $b\approx\pm2.5$, and a smaller one around $b\approx\pm4.5$. In \cref{fig:lrp_kirch_RFF_PINN_scatter_fs_2}, the largest peak is in $b\approx0$, with two lower ones around $b\approx\pm2.5$ and $b\approx\pm4$. These models have a training error of $8.99\times10^{-04}$ and $4.31\times10^{-02}$, and a validation error of $6.04\times10^{-07}$ and $1.79\times10^{-04}$, respectively. Both of them underperform the model corresponding to the best configuration shown in \cref{tab:hyperparameter_search}. They do not share peaks of attribution with the best model, and only the peak around $b\approx\pm2$ is between them. 

\begin{figure*}[!htb]
    \centering
    \subfloat[Best PINN-RFFs model configuration.]
        {\includegraphics[trim={0 1.3cm 0 0},clip,width=0.95\linewidth]{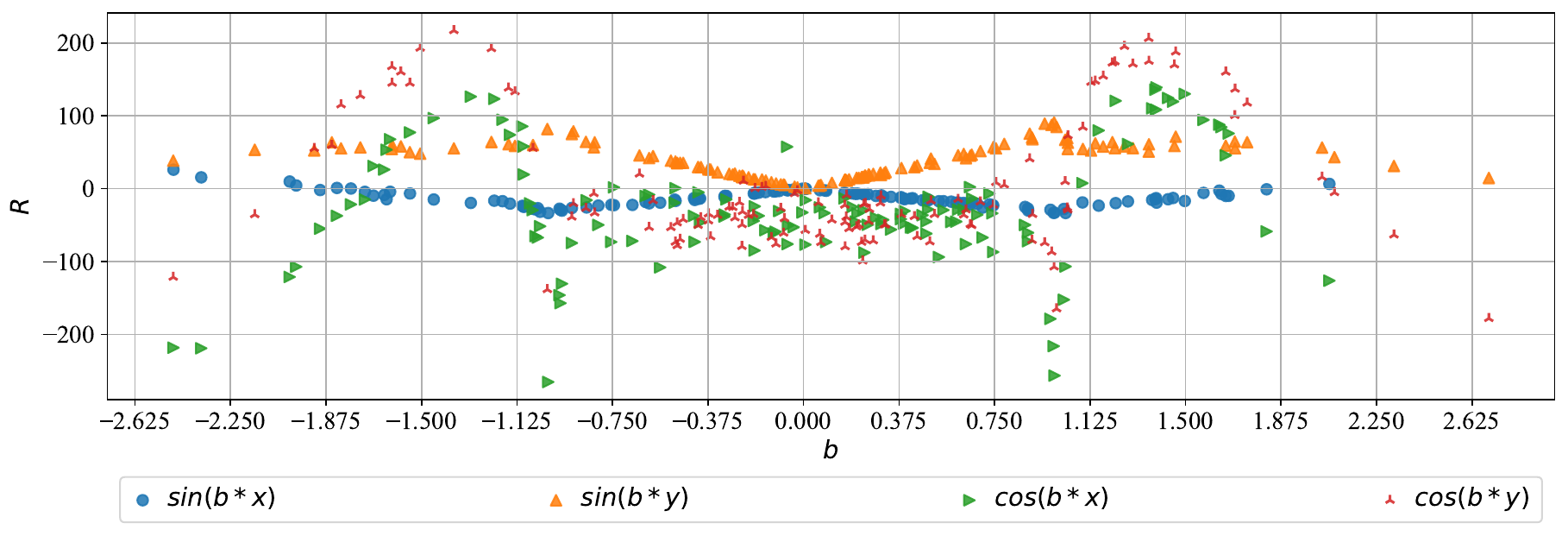}\label{fig:lrp_helm_RFF_PINN_scatter}}\\
    \subfloat[PINN-RFFs model with extended \textit{Fourier $\sigma^2_j$} = (1,2,3,4,5).]
        {\includegraphics[trim={0 1.3cm 0 0},clip,width=0.95\linewidth]{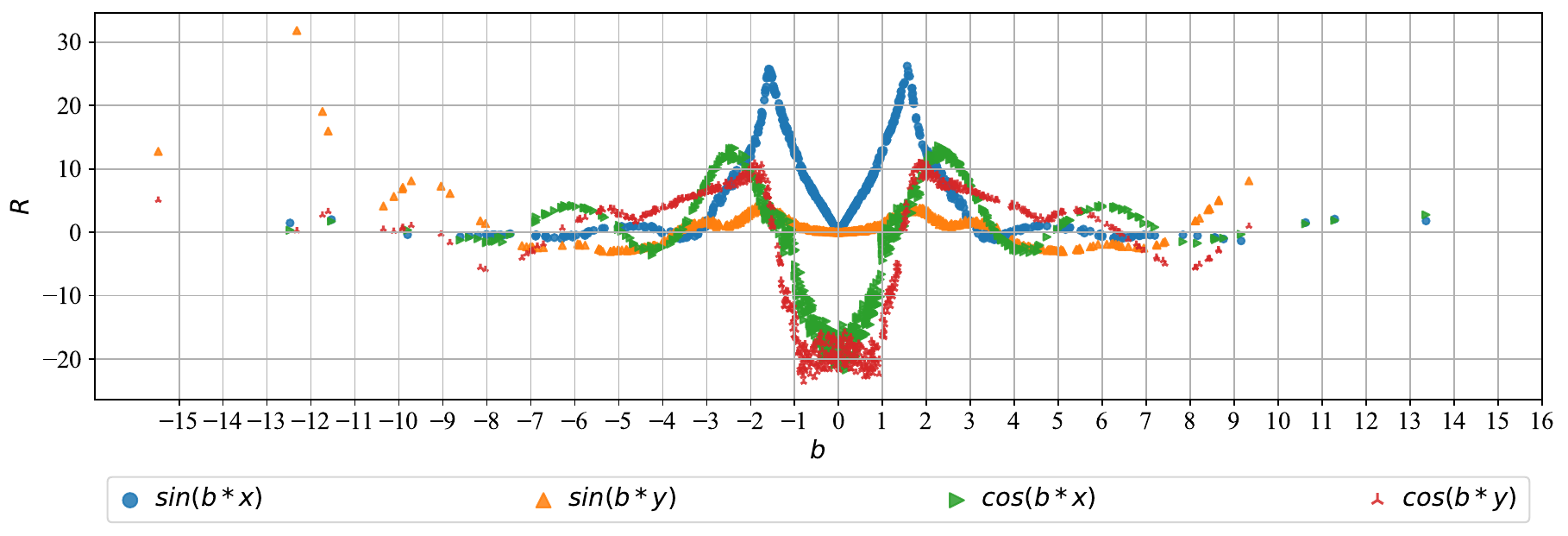}\label{fig:lrp_helm_RFF_PINN_scatter_fs_1}}\\
    \centering
    \subfloat[PINN-RFFs model with extended \textit{Fourier $\sigma^2_j$} = (1,5,10).]
        {\includegraphics[width=0.95\linewidth]{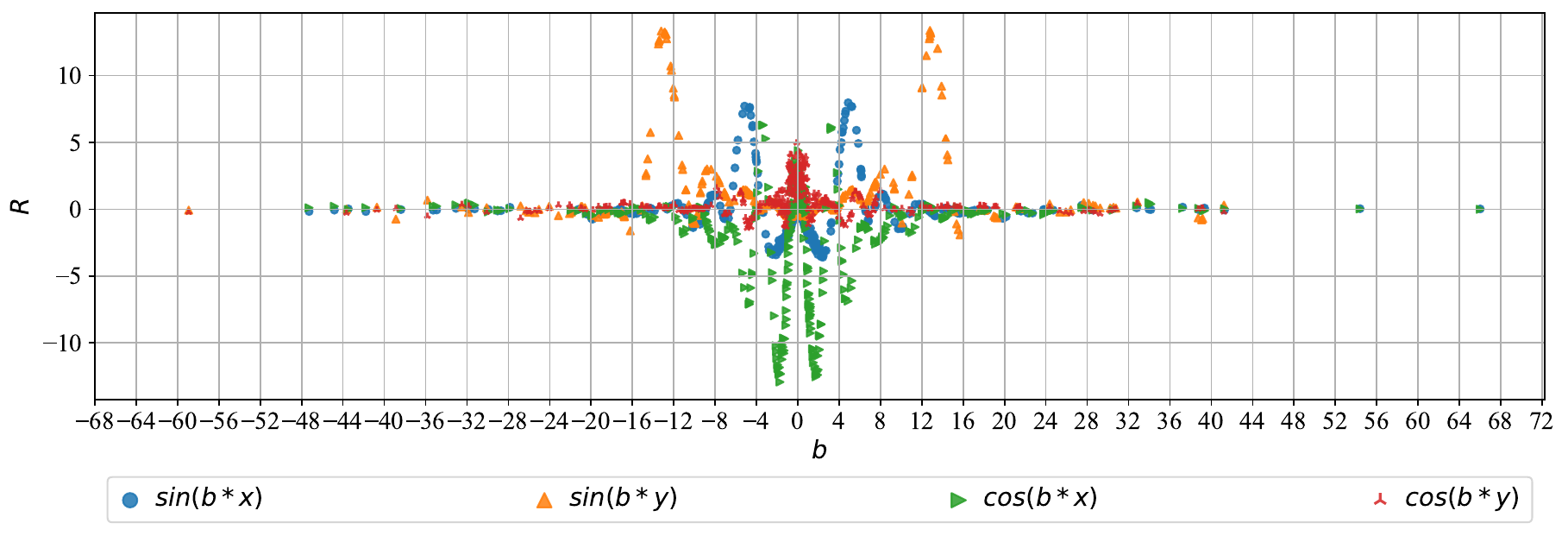}\label{fig:lrp_helm_RFF_PINN_scatter_fs_2}}\\
    \caption{Helmholtz case: contribution of RFFs to the predictions of PINNs with respect to the $\mathbf{b}$ and the type of Fourier component and spatial coordinate. No clear correlations can be found between models with different RFF, suggesting a shortage of generalizable and learnable patterns.}
    \label{fig:PINNs_RFF_helm}
\end{figure*}

The peaks shown in \cref{fig:PINNs_RFF_kirch} do not coincide with the spectral components shown in \cref{fig:kirchhoff_solution}. This shows that the models trained with RFFs do not seem to attribute a significant importance to the domain representative spectral components, and do not consistently converge to a similar learning point, from a FFs point of view. These results suggest that they tend to fall in local optima, which also fits with the performance decrease, whose contributions are scattered through different spectral components, despite the domain-specific FFs being part of the RFF space used for training. Although this model variant, namely PINN-RFFs, is not able to filter out the redundant components of the RFFs and select the ones that are representative of the problem domain, it is demonstrated that the explainability layer can be used as guidance for the feature selection process.

\begin{figure}[!htbp]
    \centering
    \subfloat[Best model configuration for the Kirhhoff problem.]{
    \includegraphics[width=0.98\textwidth]{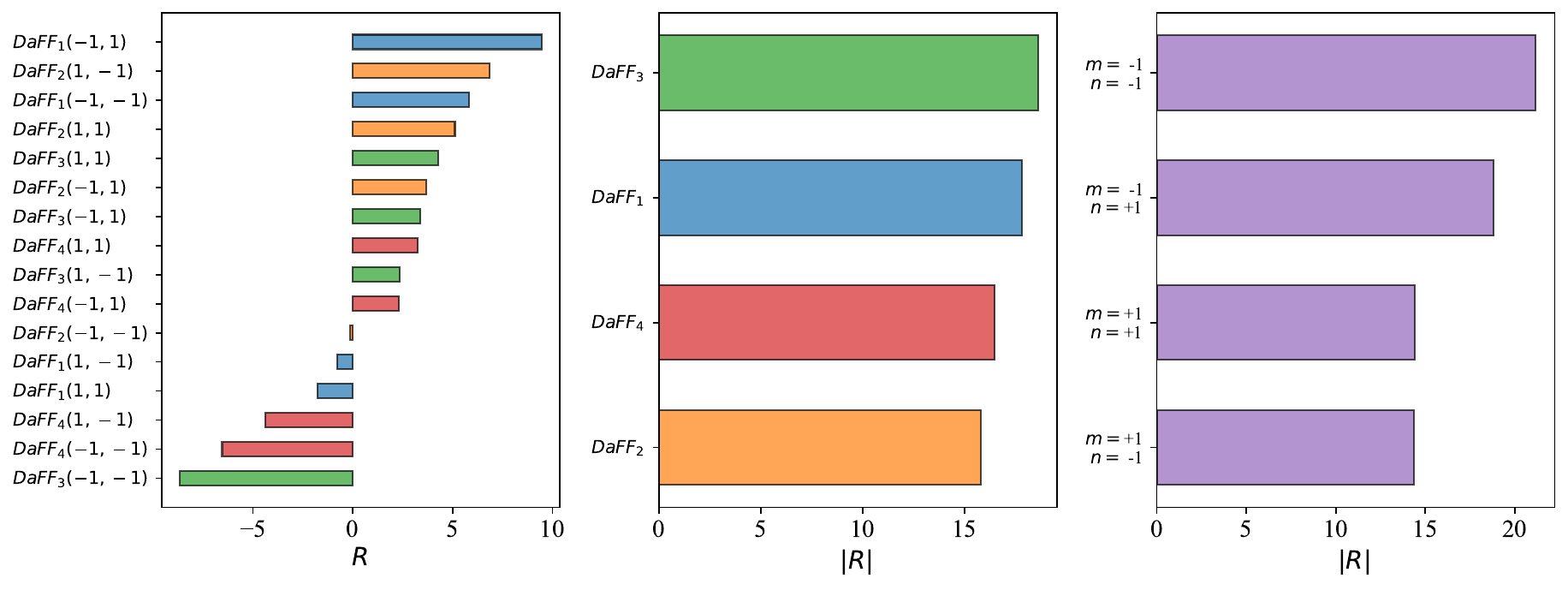}
    \label{fig:lrp_kirch_DaFF_PINN}}

    \centering
    \subfloat[Best model configuration for the Helmholtz problem.]{
    \includegraphics[width=0.98\textwidth]{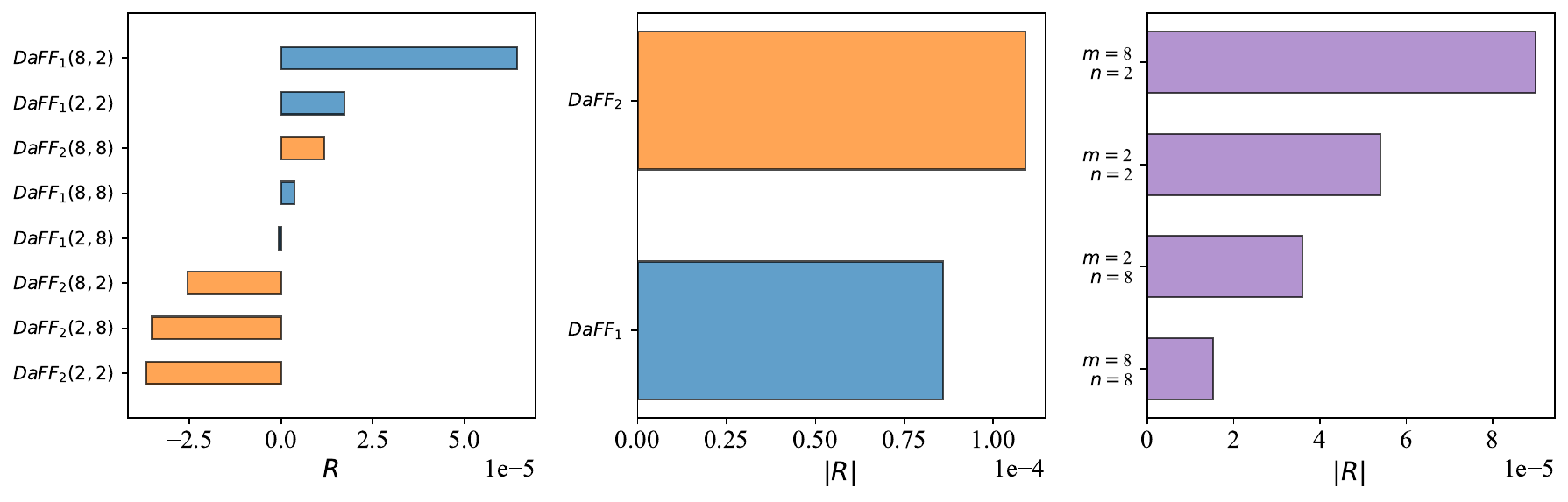}
    \label{fig:lrp_helm_DaFF_PINN}}

    \caption{Contribution of the DaFFs to the prediction of a PINN-DaFFs across the whole domain of the problems computed with LRP with respect to each DaFF, its type, and the $m_{},n_{}$ parameters.}
    \label{fig:lrp_DaFF_PINN}
\end{figure}

A similar analysis is shown in \cref{fig:PINNs_RFF_helm} for the Helmholtz problem, where the contributions of three different PINN-RFFs models are presented. The first plot, \cref{fig:lrp_helm_RFF_PINN_scatter}, corresponds to the optimal model configuration, as obtained from the hyperparameter optimization solution, while the other two are based on RFFs whose parameters $\mathbf{b}$ are sampled from a wider range. The model of \cref{fig:lrp_helm_RFF_PINN_scatter_fs_1} has a training error of $5.50\times10^{-03}$ and validation error of $7.10\times10^{-05}$, while for the model of \cref{fig:lrp_helm_RFF_PINN_scatter_fs_2} the training and validation errors are $1.39\times10^{-02}$ and $1.84\times10^{-05}$ respectively. Note that, although training error decreases, validation error is higher, which can be explained by the training error being computed based on randomly sampled collocation points, whereas the validation error is calculated in a fixed grid of the domain and the boundaries. Again, the performance of these two models is worse than the one achieved by \cref{fig:lrp_helm_RFF_PINN_scatter}, denoting local optima, and the contribution peaks do not show common spectral components that are shared by all models. Concerning the spectral components shown in the solution of \cref{fig:helmholtz_solution}, the $\mathbf{b}$ values are again different with the exception of the peak shown in \cref{fig:lrp_helm_RFF_PINN_scatter_fs_2} at $b\approx\pm13$ and some sparse points in \cref{fig:lrp_helm_RFF_PINN_scatter_fs_1} at $b\approx\pm12.5$, which are particularly close to $b \approx 4\pi$. It is interesting that, despite placing high contribution on $\mathbf{b}$ values very close to a correct spectral component, the third model delivers the worst performance, suggesting that finding the proper component is not sufficient, as the solution is influenced by the entire pool of RFFs.

Conversely, the DaFFs are by design more informative, as they are informed of the domain characteristics, such as the geometry and boundary conditions. Moreover, a much lower input dimensionality is required as the features are combined for both $x$ and $y$ coordinates. The contributions of each feature for the PINN-DaFFs model are presented in \cref{fig:lrp_DaFF_PINN} for the best performing model, as obtained from the hyperparameter optimization, whose results are summarized in \cref{tab:hyperparameter_search,tab:hyperparameter_search_helm}. For the Kirchhoff problem, it is observed that all DaFF components, i.e., for different types of DaFF and $m_{},n_{}$ values, have similar contributions. Since all the DaFF types provide similar information, it is expected that the model uses them in a similar way, as with RFFs, in which similar $\mathbf{b}$ values yield similar contributions. Interestingly, for Helmholtz, different $m_{},n_{}$ values lead to diverse contributions. For example, features with $m_{}=8, n_{}=2$ and $m_{}=2, n_{}=2$ have a larger contribution on the predictions than those with $m_{}=8, n_{}=8$.

\begin{figure}[!htbp]
    \centering
    \includegraphics[width=0.98\textwidth]{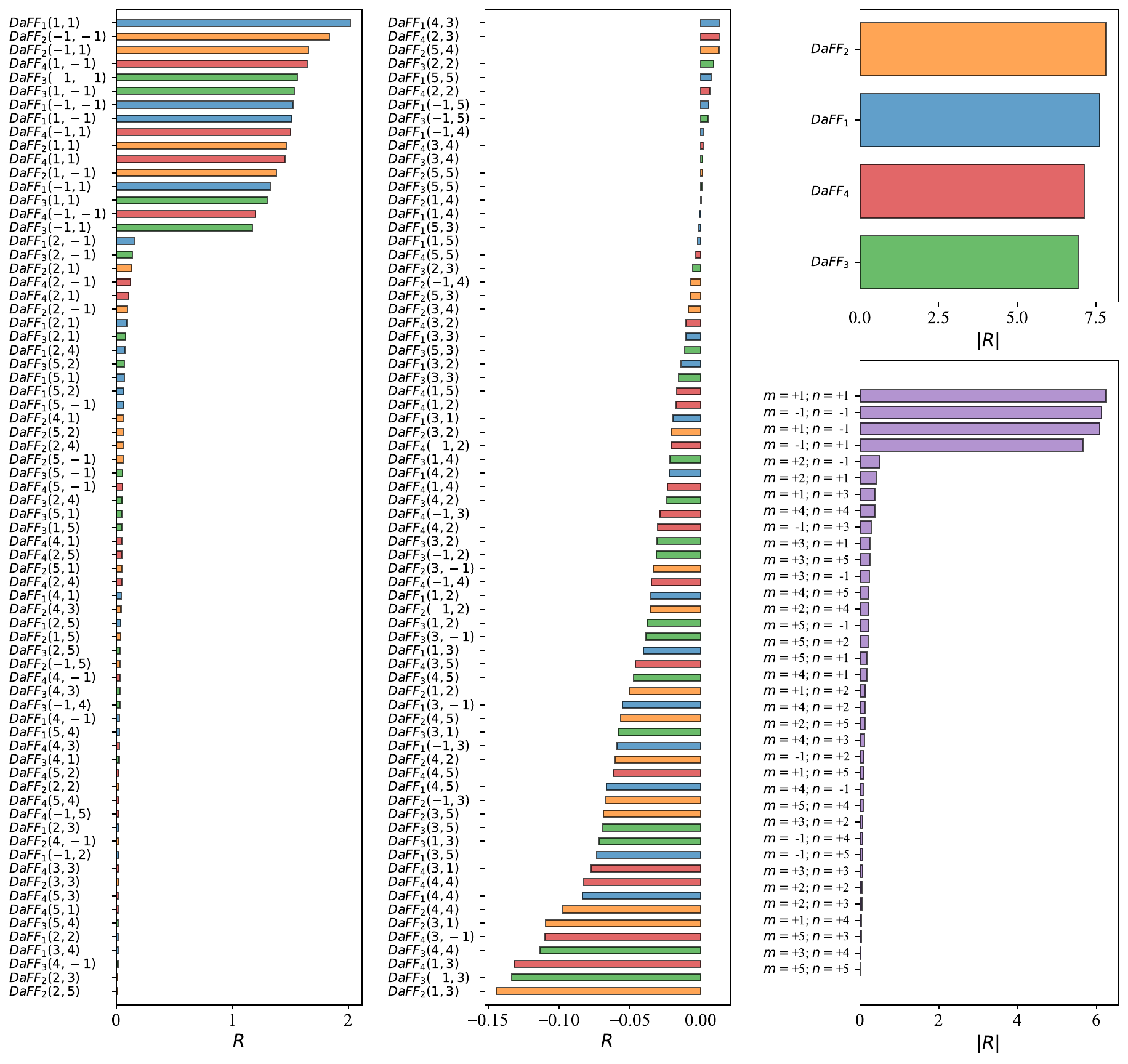}
    \caption{Contribution of DaFFs to the prediction of a PINN with additional combinations of $m_{},n_{}$ values.}
    \label{fig:lrp_kirch_DaFF_PINN_feat_select}
\end{figure}

In analogy to the $\mathbf{b}$ values of the PINN-RFFs models, a wider range of harmonic indices $m_{},n_{}$ is utilized in order to explore the learned physics of the PINN-DaFFs model. A PINN-DaFFs model, as presented in \cref{tab:hyperparameter_search}, with \textit{DaFF $(m,n)$}, where  $m, n \in \{-1,1,2,3,4,5\}$, results in a training loss of $1.60\times10^{-03}$, which corresponds to a validation loss equal to $8.60\times10^{-16}$, computed as the data MSE. The input contributions of this model can be seen in \cref{fig:lrp_kirch_DaFF_PINN_feat_select}. Only the contributions aggregated by $(m,n)$ values, right bottom figure, show how lower $(m,n)$ values are more important for the predictions, which align with the values for the best performing models. 

\begin{figure}[!htbp]
    \centering
    \includegraphics[width=0.98\textwidth]{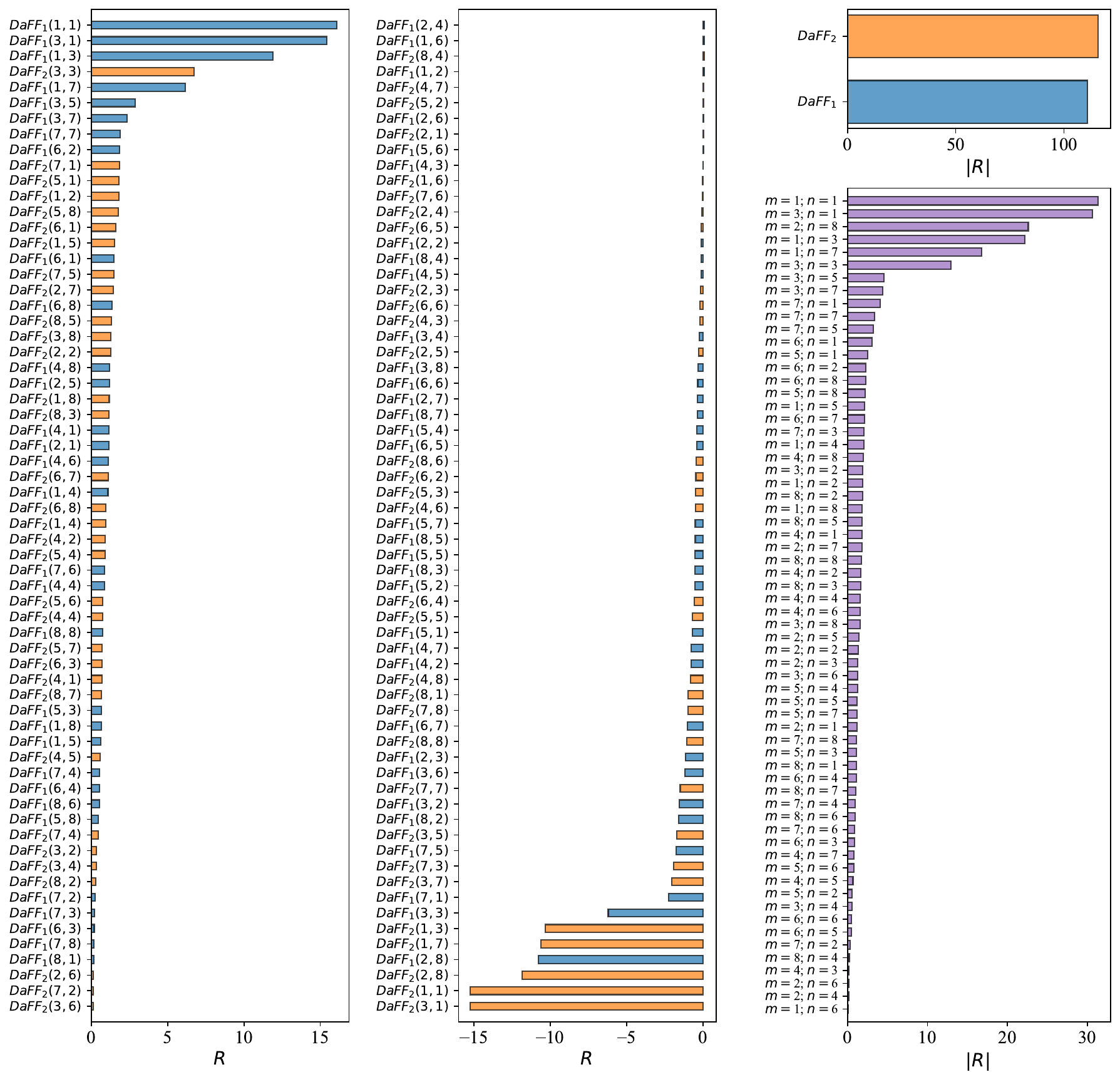}
    \caption{Contribution of DaFFs to the prediction of a PINN-DaFFs with more combinations of $m_{},n_{}$ values.}
    \label{fig:lrp_helm_DaFF_PINN_feat_select}
\end{figure}

The results of the Helmholtz problem delivered by the PINN-DaFFs model, which was configured with the harmonic indices $m,n \in \{1,2,3,4,5,6,7,8\}$, are presented in \cref{fig:lrp_helm_DaFF_PINN_feat_select}. The results are consistent with those reported for the Kirchhoff problem: a higher training error of $2.10\times10^{-04}$ due to the increased dimensionality of the model, which is owed to the redundancy of the FFs, but a very low validation error of $1.83\times10^{-10}$. Similarly, all the DaFFs have comparable contributions, while some combinations of harmonic indices $m_{},n_{}$ are significantly more important than others. In contrast to the Kirchhoff case, there are several $m_{},n_{}$ values for the Helmholtz problem that are not present in the optimal model, such as $m_{}=1,n_{}=1$ or $m_{}=3,n_{}=1$. This is a symptom of the model having found a different approximation than the best PINN-DaFFs configuration shown before, and showcases one of the challenges of using relevance attribution scores as guiding tools for feature selection: if the models find a suboptimal approximation, then the attributions will, most likely, point in the direction of the found local optima.

\section{Conclusion}
\label{sec:conclusion}

In this work, an extension of PINN models is proposed with DaFFs, which encode input features. The proposed DaFFs are extracted from the eigenvalue problem of the Laplace operator, defined on the domain of the PDE to be solved. By definition, these features are designed to satisfy the boundary and initial conditions, thus requiring only the PDE residuals for the model training. This, in return, removes one of the well-known pathologies related to vanilla PINNS, which is related to the gradient stiffness and is so far treated by means of loss balancing techniques. The proposed PINN-DaFFs model is benchmarked against two well-established PINN models, namely vanilla PINNs and PINN-RFFs. Within this context, two well-known benchmark problems are considered, which are commonly encountered in physics and engineering applications. It is demonstrated that the use of DaFFs for the encoding of the PINN inputs takes into account the geometrical and physical characteristics of the domain, resulting in an increased model expressiveness.

Furthermore, an LRP-based framework is proposed in order to address the black-box nature of PINNs and provide a critical comparison of the proposed model with the state-of-the-art from an explainability point of view. Under this perspective, it is observed that vanilla PINNs tend to have a bias towards certain collocation points, which showcases the challenge of interpreting a model whose inputs are merely spatial coordinates. On the other hand, it is shown that PINN-RFFs models are characterized by a scattering of the contributions across all the FFs. Although the vast majority of contribution scores are concentrated around certain spectral areas, their position is randomly selected for each training, without offering further insights and interpretation to the learned physics. Lastly, the proposed PINN-DaFFs variant results in comparable contributions for each DaFF, which is expected as these are tailored to the geometrical characteristics of the domain, including the physics at the boundaries. Although both types of FFs allow for further interpretation of the learned physics, the use of DaFFs results in a more expressive representation, where a clear threshold can be used in combination with the LRP as a means of feature selection in the FFs space. 

We also show how models with a wider range of RFFs $\mathbf{b}$ values or DaFFs $m_{},n_{}$ hyperparameters can be tuned using LRP, in order to choose the most performant combination of features. Although the approach can simplify the tuning of hyperparameters, it should be noted that the reliability of the guidance offered by LRP, as well as other feature attribution score XAI methods, is linked to the performance of the model. As such, the input contributions extracted from models with poor performance will typically not be relevant for training high-performance models.

Future work aims to generalize the proposed modeling approach to problems with more complex physics and non-homogeneous boundary conditions. Furthermore, the proposed framework may also be of interest for more challenging structural-mechanics settings, such as geometrically nonlinear plates and shell structures. Extending the present formulation to such cases would require adapting the feature construction to more complex operators, geometries, and boundary conditions, and is therefore left for future work. Additionally, the use of DaFFs will be explored as a means of learning on the basis of sparse observations for unknown or partly known systems \cite{Tatsis2022}. Lastly, the application of XAI methods such as LRP can be used to reveal biases or, more generally, provide insights and interpretation of the inner workings of PINN-based models. A better understanding of the model mechanics can lay the ground for insightful and physics-aware developments, as well as for solutions of existing issues associated with such models.

\section*{Acknowledgements}
This work is partly funded by the Research Council of Norway through the EXAIGON project (project no. 304843).

\bibliographystyle{elsarticle-num}

\bibliography{references,references_al}

\end{document}